\documentclass[twocolumn]{article}

\usepackage{geometry}
\usepackage{amsmath}
\usepackage{amsthm}
\usepackage{amssymb}

\usepackage[utf8]{inputenc}
\usepackage{hyperref}

\usepackage[sort&compress,numbers,square]{natbib}
\usepackage{graphicx, color}

\usepackage{mathrsfs} 

\usepackage{adjustbox}
\usepackage{multirow}
\usepackage{subcaption}
\usepackage{array}
\usepackage[font=small,skip=1pt]{caption}
\newcolumntype{C}[1]{>{\centering\let\newline\\\arraybackslash\hspace{0pt}}m{#1}}

\title{Self-Attention Dense Depth Estimation Network for Unrectified Video Sequences}
\author{Alwyn Mathew \and Aditya Prakash Patra \and Jimson Mathew}
\date{
	Indian Institute of Technology Patna \\ \texttt{\{alwyn.pcs16, aditya.cs16, jimson\}@iitp.ac.in}\\[2ex]%
}

\begin{document}
	
\setlength{\abovedisplayskip}{0pt}
\setlength{\belowdisplayskip}{0pt}

\maketitle
	
\begin{abstract}
	The dense depth estimation of a 3D scene has numerous applications, mainly in robotics and surveillance. LiDAR and radar sensors are the hardware solution for real-time depth estimation, but these sensors produce sparse depth maps and are sometimes unreliable. In recent years research aimed at tackling depth estimation using single 2D image has received a lot of attention. The deep learning based self-supervised depth estimation methods from the rectified stereo and monocular video frames have shown promising results. We propose a self-attention based depth and ego-motion network for unrectified images. We also introduce non-differentiable distortion of the camera into the training pipeline. Our approach performs competitively when compared to other established approaches that used rectified images for depth estimation.
	
\end{abstract}

\let\thefootnote\relax\footnote{\copyright 2020 IEEE.  Personal use of this material is permitted.  Permission from IEEE must be obtained for all other uses, in any current or future media, including reprinting/republishing this material for advertising or promotional purposes, creating new collective works, for resale or redistribution to servers or lists, or reuse of any copyrighted component of this work in other works.}

\section{Introduction}
\label{sec:intro}
 
Several prior works for depth estimation from monocular images doesn't work with raw images straight out of the camera, which would be distorted and unrectified. Instead, these raw images are corrected with image processing techniques before being fed into the network to predict an undistorted rectified depth map. We propose a novel fully differentiable architecture which allows estimating the distorted unrectified depth directly from a raw image without the need for any pre-processing techniques to correct the image. This pipeline in turn saves time and computation power, and can be used directly in real-time scenarios without any prerequisites. The parameters which define distortion vary slightly with the environment in which the camera is present. Our model exploits the fact that more or less the amount of distortion in an image is fixed as long as we use the same camera. This allows us to pre-define a transformation flowfield for undistorting the image and using it in the training pipeline. Image rectification on-the-fly with prerecorded camera parameters has an overhead of around 75msec for a typical 30 FPS (frames per second) camera in one second.
Our proposed model for unrectified distorted images doesn't have any overhead at inference when compared to models based on rectified images as it uses a neural network with the same number of trainable parameters. Image rectification also leads to information loss from the raw image. Higher the camera distortion, the higher the loss of pixel information from the raw data. In the KITTI dataset due to the distortion effect of the camera, the images have been rectified and cropped to 1242 $\times$ 375 such that the size of the rectified undistorted images is smaller than the raw unrectified distorted image size of 1392 $\times$ 512 pixels. Image rectification in KITTI resulted in the pixel loss of 10.77\% and 27.34\% along the width and height of the image, respectively. This pixel loss becomes more prominent at higher resolutions. We address the above issues by proposing the following contributions:

\begin{figure}
  \centering
  \resizebox{0.5\textwidth}{!}{
  \newcommand{\turnheightnew}{0.25\columnwidth}
\centering

\begin{tabular}{@{\hskip 0.5mm}c@{\hskip 0.5mm}c@{\hskip 0.5mm}c@{\hskip 0.5mm}c@{\hskip 0.5mm}c@{}}

{\hspace{6mm} {\Large Unrectified input image}} &
{\hspace{6mm} {\Large Attention map}} \\

\includegraphics[height=\turnheightnew]{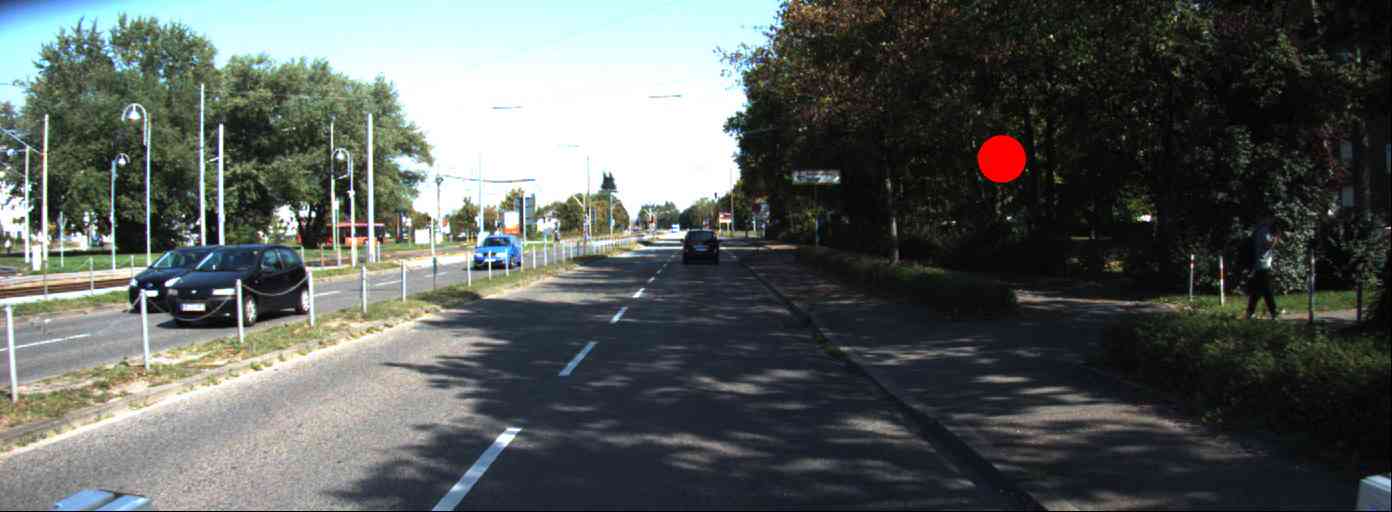} &
\includegraphics[height=\turnheightnew]{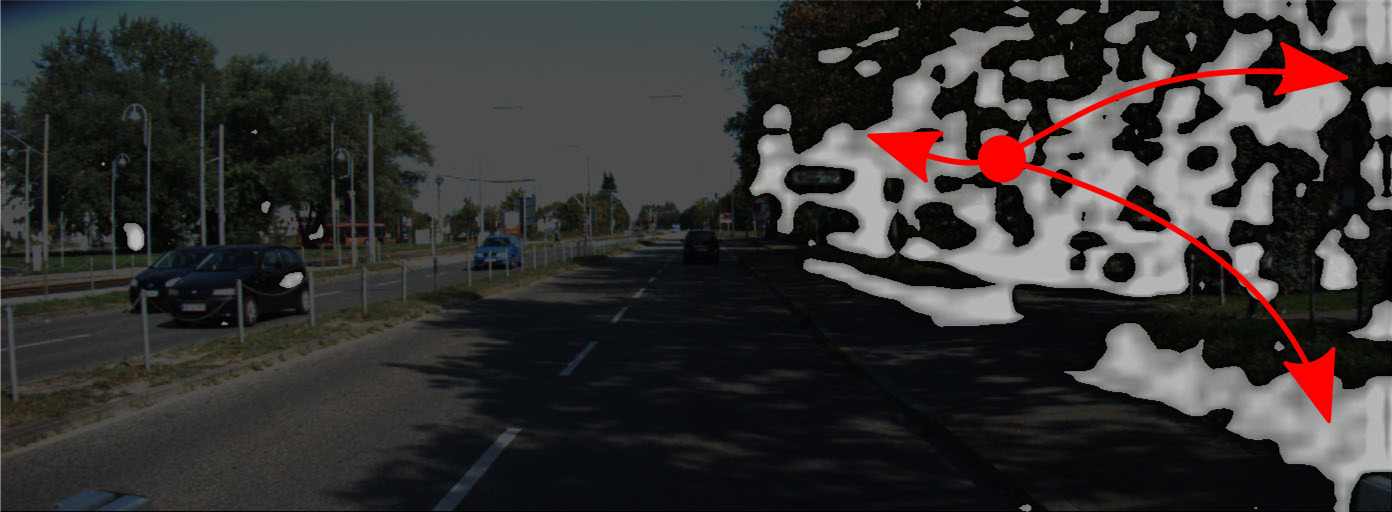} \\

\includegraphics[height=\turnheightnew]{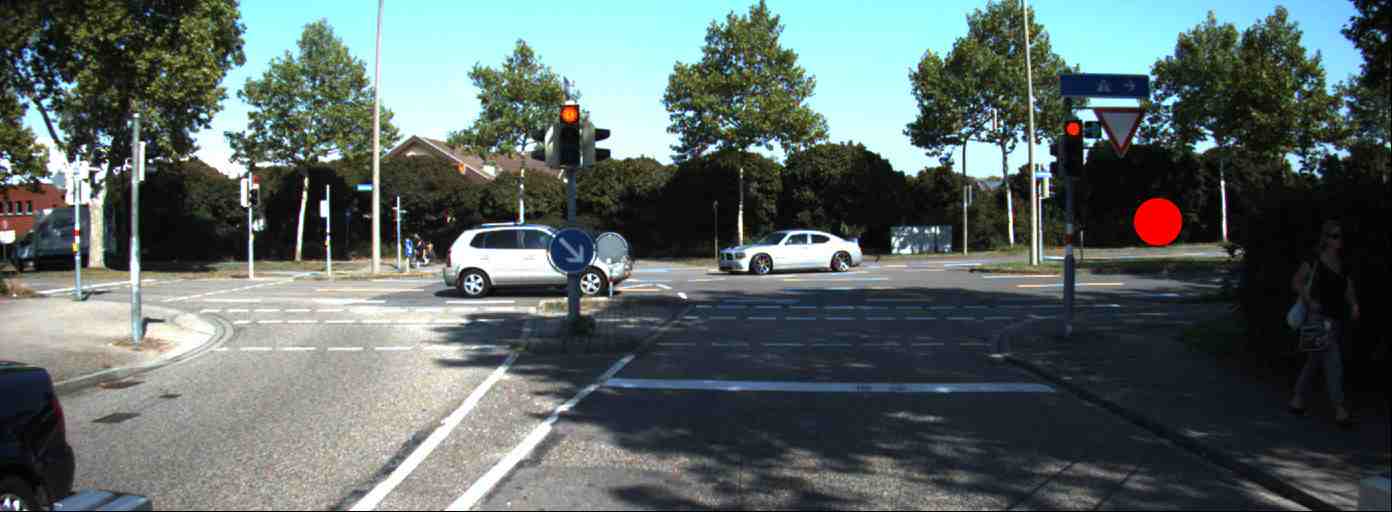} &
\includegraphics[height=\turnheightnew]{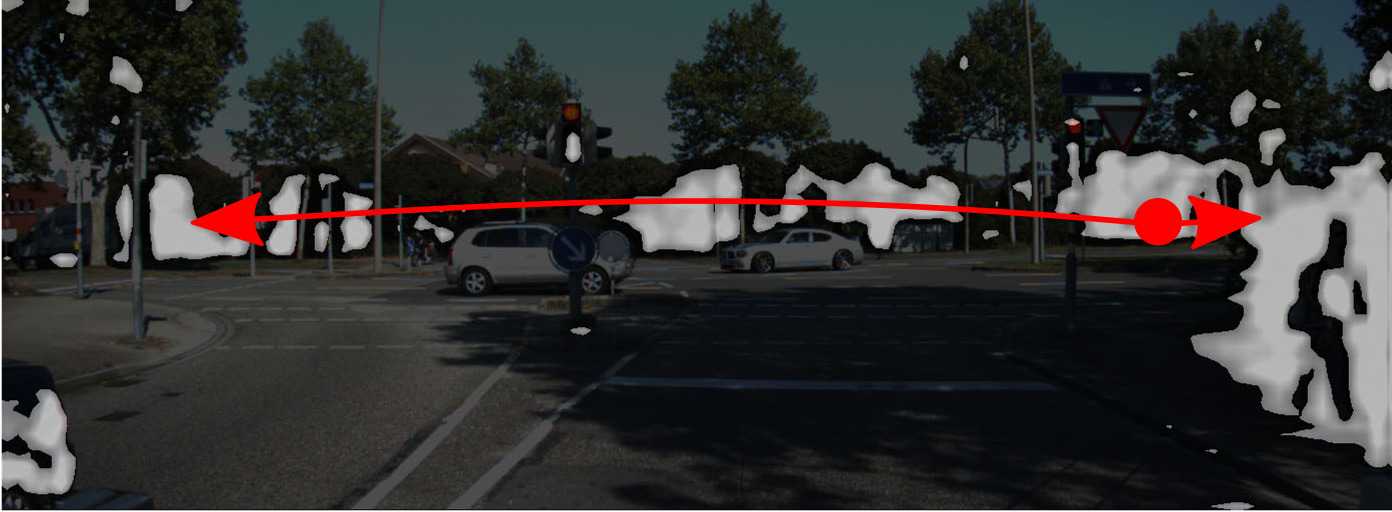} \\

\end{tabular}
}
  \caption{Visualization of self-attention map. Left: Shows Query locations marked with a red dot on the unrectified input image. Right: Attention map for those query locations overlayed on the unrectified input image with arrows summarizing most-attended regions. Rather than attending to spatially adjacent pixels, the proposed network learns to attended pixels with similar color and texture.
  }
  \label{fig:att}
\end{figure}

\begin{enumerate}
\itemsep -0.5em 
\item Introduced an end to end fully differentiable novel pipeline to tackle monocular depth estimation on distorted and unrectified images.
\item Proposed a novel self-attention depth network to handle long range dependencies in the feature map leading to sharper depth maps.
\item Incorporated instance normalisation throughout model architecture to handle style variation in the input data.
\end{enumerate}

Some of the prior self-supervised monocular depth estimation works are briefed here.
Zhou \textit{et al}., \cite{zhou2017unsupervised} proposed an unsupervised depth estimation end-to-end framework from video sequence where view synthesis act as the supervisory signal.
Mahjourian \textit{et al}., \cite{mahjourian2018unsupervised} introduced a 3D loss to enforce consistency of the estimated 3D point cloud.
Yin \textit{et al}., \cite{yin2018geonet} jointly learned monocular depth, optical flow and ego motion from videos in unsupervised manner.
Luo \textit{et al}., \cite{luo2018every} also jointly learned monocular depth, optical flow and ego motion from video but used these result to produce motion mask, occlusion mask, 3D motion map for rigid background and dynamic objects.
Godard \textit{et al}., \cite{godard2019digging} achieved state of the art results in KITTI rectified dataset with minimum reprojection loss, full-resolution mutli-scale sampling and a mask that ignores training pixels that violate camera motion assumptions.

\setlength{\belowcaptionskip}{-1pt}
\begin{figure*}[ht]
	\centering
	\begin{subfigure}{.45\textwidth}
		\centering
		\includegraphics[width=1\linewidth]{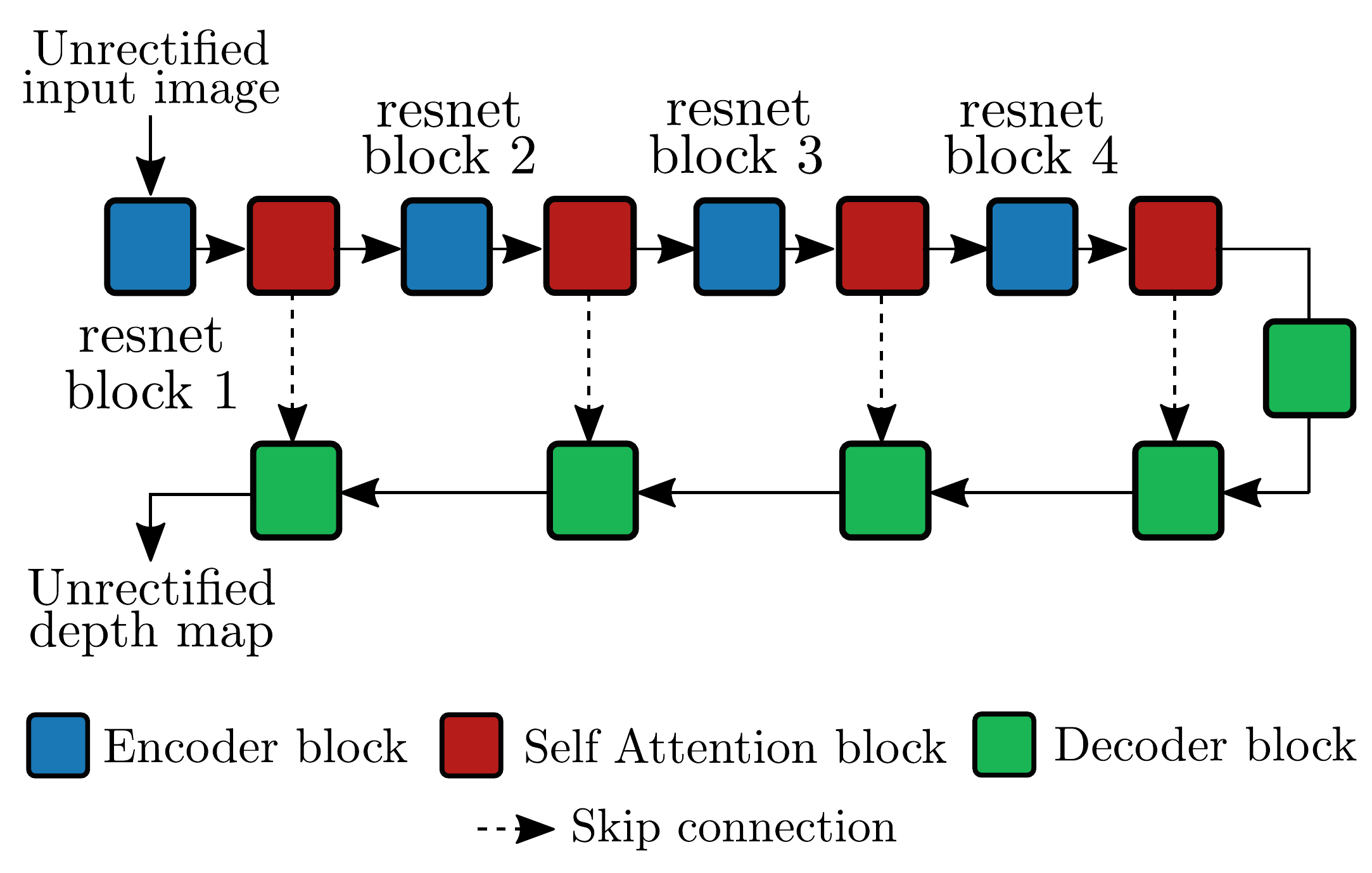}
		\caption{}
		\label{fig:input}
	\end{subfigure}
	\begin{subfigure}{.45\textwidth}
		\centering
		\includegraphics[width=1\linewidth]{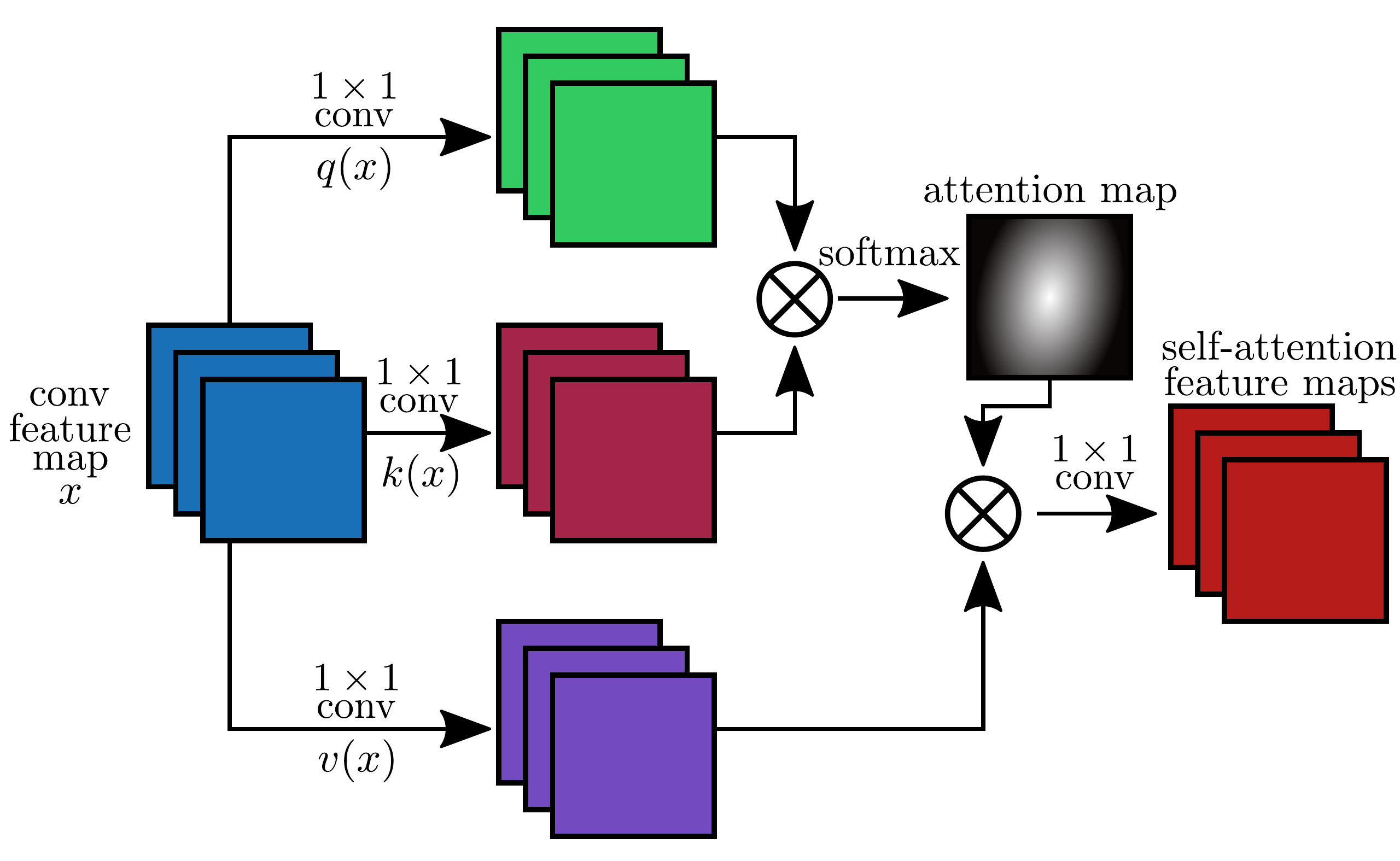}
		\caption{}
		\label{fig:output}
	\end{subfigure}
	\caption{(a) Proposed self-attention depth estimation network. (b) Self-attention block. $\bigotimes$ denote matrix multiplication.}
	\label{fig:net}
\end{figure*}

\section{Synthesis of unrectified image}

All the prior works on monocular depth estimation have used rectified images during training to estimate depth. But the raw images from camera are unrectified images; henceforth we enable an efficient way to train the depth estimation network on unrectified images. Rectification is required on images like stereo pairs to rectify the error in rotation and translation between cameras. Image distortion is prevalent in images caused by varying magnification in relation to the angular distance to the optical axis. Due to the imperfect alignment of the lens or sensor, distortion may get decentered, but that is rarely an issue in current cameras. Usually, these images are fixed beforehand, as it would affect the projection geometry from one view to another. Distortion of a typical $90^\circ$ FOV (field of view) camera can be modeled with few intrinsic parameters where $k_1,k_2,k_3,p_1,p_2$ are the known distortion parameters. We can obtain the undistorted image coordinate $\hat{c}_{t-1}^{undist}$  at time $t-1$ from distorted image coordinate $c^{dist}_{t}$ at time $t$ by Equ.~\ref{eq:1}.

\begin{align} \label{eq:1}
\hat{c}_{t-1}^{undist} \approx \mathbf{K} \mathbf{T}_{t \rightarrow t-1}D_{t}\mathbf{K}^{-1}\xi\{c^{dist}_{t}\}
\end{align}

where $\mathbf{K}$ is the camera intrinsic $3 \times 3$ matrix that transforms camera coordinates to image coordinates which comprises of focal length $f_x, f_y$, principal point offset $x_0, y_0$ along x and y directions and axis skew $s$ as shown in Equ.~\ref{eq:2}, $\mathbf{T}_{t \rightarrow t-1}$ camera transformation $3 \times 4$ matrix from target view frame at $t$ to source view frame at $t-1$ consisting of $3 \times 3$ rotation matrix $\mathbf{R}$ and $3 \times 1$ translation vector $\mathbf{t}$ as shown in Equ.~\ref{eq:3}, $D_t$ is the per-pixel depth  predicted by the network and $\xi$ is the undistortion function with pre-computed distortion parameters.

\begin{align} \label{eq:2}
\mathbf{K} =
\left[ \begin{array}{ccc} 
f_x & s & x_0\\
0 & f_y & y_0\\
0 & 0 & 1
\end{array} \right]
\end{align}

\begin{align} \label{eq:3}
\mathbf{T} = [ \mathbf{R} \, |\, \mathbf{t}] = 
\left[ \begin{array}{ccc|c} 
r_{1,1} & r_{1,2} & r_{1,3} & t_1 \\
r_{2,1} & r_{2,2} & r_{2,3} & t_2 \\
r_{3,1} & r_{3,2} & r_{3,3} & t_3
\end{array} \right]
\end{align}

Per-pixel depth $D_t$ of frame at time $t$ and camera transformation $\mathbf{T}_{t \rightarrow t-1}$ from frame at time $t$ to $t-1$ are estimated by self-attention depth and pose network, respectively as shown in Fig.~\ref{fig:net}. Pose network \cite{zhou2017unsupervised} takes three adjacent distorted image frames at time $t-1$, $t$, $t+1$ and predicts camera transformation $\mathbf{T}_{t \rightarrow t-1}$ and $\mathbf{T}_{t \rightarrow t+1}$. While self-attention  depth network takes single distorted image frame $I^{dist}_{t}$ at time $t$ to predict per-pixel depth $D_t$. During inference, only self-attention depth network is used to predict depthmap of an input image.

\subsection{Image coordinates to World coordinates} \label{sec:2.1}

With intrinsic matrix $\mathbf{K}$, estimated depth $D_t$ and undistortion function $\xi$, we can project distorted image coordinates to world coordinates as shown in Equ.~\ref{eq:4}.

\begin{align} \label{eq:4}
\left[\begin{array}{c} 
X \\
Y \\
Z
\end{array}\right]
\approx D_{t}\mathbf{K}^{-1}\xi \Bigg\{
\left[\begin{array}{c} 
x^{dist}_t \\
y^{dist}_t \\
1
\end{array}\right]\Bigg\}
\end{align}

where $(x^{dist}_t, y^{dist}_t)$ is unrectified image coordinates and $(X, Y, Z)$ is world coordinates. Undistortion function $\xi$ is used to pre-compute undistortion map which is used in the training pipeline as shown in Equ.~\ref{eq:5}.

\begin{align} \label{eq:5}
\left[\begin{array}{c} 
x^{undist}_t \\
y^{undist}_t \\
\end{array}\right]
\approx
\xi \Bigg\{
\left[\begin{array}{c} 
x^{dist}_t \\
y^{dist}_t \\
\end{array}\right]\Bigg\}
\end{align}

The mathematical formulation of radial and tangential distortions used in the undistortion function $\xi$ are shown below in Equ.~\ref{eq:6} and Equ.~\ref{eq:7}

\begin{align} \label{eq:6}
\begin{gathered}
\hat{x}^{dist}_t \approx x^{undist}_t(1+k_1r^2+k_2r^4+k_3r^6) \\
\hat{y}^{dist}_t \approx y^{undist}_t(1+k_1r^2+k_2r^4+k_3r^6)
\end{gathered}
\end{align} 

Here $x^{undist}_t$ and $y^{undist}_t$ denotes undistorted pixel coordinates, $\hat{x}^{dist}_t$, $\hat{y}^{dist}_t$ denotes radially distorted pixel coordinates and $r^2={x^{undist}_t}^2+{x^{undist}_t}^2$.
Tangential distortion caused by lens misalignment can be formulated with distortion parameters $p_1$ and $p_2$ along with radial distortion as shown in Equ.~\ref{eq:7}.

\begin{align} \label{eq:7}
\begin{gathered}
x^{dist}_t \approx \hat{x}^{dist}_t + 2p_1x^{undist}_ty^{undist}_t + p_2(r^2+2{x^{undist}_t}^2) \\
y^{dist}_t \approx \hat{y}^{dist}_t + p_1(r^2+{y^{undist}_t}^2) + 2p_2x^{undist}_ty^{undist}_t
\end{gathered}
\end{align}

$x^{dist}_t$ and $y^{dist}_t$ are resultant image coordinate after radial and tangential distortion. Distortion mapping in Equ.\ref{eq:5} is used to inverse map distorted image coordinates to undistorted image coordinates. These undistorted image coordinates can be projected to world coordinates by dot operation with inverse camera intrinsic $\mathbf{K}^{-1}$ and estimated per-pixel depth $D_t$ as shown in Equ.~\ref{eq:8}.

\begin{align} \label{eq:8}
\left[\begin{array}{c} 
X \\
Y \\
Z
\end{array}\right]
\approx D_{t}\mathbf{K}^{-1}
\left[\begin{array}{c} 
x^{undist}_t \\
y^{undist}_t \\
1
\end{array}\right]
\end{align}

\subsection{World coordinates to Image coordinates} \label{sec:2.2}

The estimated transformation matrix $\mathbf{T}_{t \rightarrow t-1}$ from image frame at time $t$ to $t-1$ along with intrinsic matrix $\mathbf{K}$ is used to transform world coordinates to image coordinates of frame at time $t-1$ as shown in Equ.~\ref{eq:9}. 

\begin{align} \label{eq:9}
\left[\begin{array}{c} 
x^{undist}_{t-1} \\
y^{undist}_{t-1} \\
1
\end{array}\right]
=
\left[\begin{array}{c} 
x/z \\
y/z \\
1
\end{array}\right]
\equiv
\left[\begin{array}{c} 
x \\
y \\
z
\end{array}\right]
\approx \mathbf{K}T_{t \rightarrow t-1}
\left[\begin{array}{c} 
X \\
Y \\
Z \\
1
\end{array}\right]
\end{align}

\subsection{View Synthesis}

The image coordinate projection from frame $t$ to $t-1$ as discussed in subsection \ref{sec:2.1} and \ref{sec:2.2} can only project distorted image coordinates at $t$ frame to $t-1$ unrectified image coordinates as shown in Equ.~\ref{eq:1}. For view synthesis, we propose a trick with image undistortion function $\varphi$ using the pre-computed distortion map to undistort the source image $I^{dist}_{t-1}$ at t-1 as shown in Equ.~\ref{eq:10}. It should be noted that this undistortion operation is performed only in the training phase.

\begin{align} \label{eq:10}
I^{undist}_{t-1} = \varphi \{ I^{dist}_{t-1} \}
\end{align}

This unrectified image $I^{undist}_{t-1}$ can be used to sample pixel values according to projected image unrectified coordinates $x^{undist}_{t-1}, y^{undist}_{t-1}$ to synthesize unrectified image $\hat{I}^{dist}_{t}$ at time $t$ as shown in Equ.~\ref{eq:11}. $\big \langle \big \rangle$ is the bilinear sampler. For simplicity, synthesis of distorted image $\hat{I}^{dist}_{t}$ at time $t$ from distorted image $I^{dist}_{t-1}$  at time $t-1$ is shown. But the proposed method synthesis $I^{dist}_{t}$ from images $I^{dist}_{t+1}$  at time $t+1$ too.

\begin{align} \label{eq:11}
\hat{I}^{dist}_{t} = I^{undist}_{t-1} \big \langle x^{undist}_{t-1}, y^{undist}_{t-1} \big \rangle
\end{align}

\section{Reconstruction loss}

The depth network takes one distorted unrectified image $I^{dist}_{t}$ at time $t$ to reconstruct $\hat{I}^{dist}_{t}$ by inverse wrapping nearby distorted views $I^{dist}_{t-1}$ and $I^{dist}_{t+1}$ with estimated depth $D_{t}$. The reconstruction loss $L^s$ at scale $s$ of our network consist of two parts, photometric loss $L_{p}$ and smoothness loss $L_s$. We follow \cite{godard2017unsupervised} in using mixture of L1 and SSIM $S(.)$ with $\alpha_1=0.85$ and $\alpha_2=0.15$ for the reconstruction loss for supervising the networks as shown in Equ.~\ref{eq:12}.

\begin{align} \label{eq:12}
L_{p}^{t-1} = \eta_{t-1}\Bigg(\alpha_1 \frac{1-S(I_t^{dist},\hat{I}_{t}^{dist})}{2} + \alpha_2 |I_t^{dist}-\hat{I}_{t}^{dist}|\Bigg)
\end{align}

Here $\eta_{t-1}$ stands for auto mask \cite{godard2019digging} which is used to mask out the pixels which do not change across time steps. This mask allows to remove static objets like a car moving at same speed as the camera and far away objects like sky, since they don't contribute of the loss. We also use an edge-aware smoothness loss is as shown in Equ.~\ref{eq:13}.

\begin{align} \label{eq:13}
L_{s}^{t-1} = |\bigtriangledown_u D_{t}| e^{-|\bigtriangledown_u I_{t}^{dist}|} + |\bigtriangledown_v  D_{t}| e^{-|\bigtriangledown_v I_{t}^{dist}|}
\end{align}

\begin{table*}[t]
\vspace{2mm}
\centering
\resizebox{\textwidth}{!}{
\begin{tabular}{l|c|cccc|ccc}
    Approach & Input &\multicolumn{4}{c|}{Lower is better} & \multicolumn{3}{c}{Higher is better}\\
    \cline{3-9}
    & & Abs Rel & Sq Rel & RMSE &  RMSE log & $\delta < 1.25$ & $\delta < 1.25^2$ & $\delta < 1.25^3$ \\
    \hline
    Zhou~\cite{zhou2018unsupervised}                  & rectified & 0.176 & 1.532 & 6.129 & 0.244 & 0.758 & 0.921 & 0.971 \\
    Mahjourian~\cite{mahjourian2018unsupervised}      & rectified & 0.134 & 0.983 & 5.501 & 0.203 & 0.827 & 0.921 & 0.981 \\
    GeoNet~\cite{yin2018geonet}                       & rectified & 0.132 & 0.994 & 5.240 & 0.193 & 0.833 & 0.953 & 0.985 \\
    EPC++~\cite{luo2018every}                         & rectified & 0.120 & 0.789 & 4.755 & 0.177 & 0.856 & 0.961 & 0.987 \\
    Monodepth2~\cite{godard2019digging}              & rectified & 0.090 & 0.545 & 3.942 & 0.137 & 0.914 & 0.983 & 0.995 \\
    \hline
    Our (resnet18 w/o attention) & unrectified & 0.1481 & 1.2579 & 5.805  & 0.221 & 0.817 & 0.941 & 0.978 \\
    Our (resnet50 w/o attention) & unrectified & 0.1254 & 1.0812 & 5.407  & 0.197 & 0.866 & 0.956 & 0.983 \\
    \hline
    Our (resnet18 w/ attention BN) & unrectified & 0.1305 & 1.0462 & 5.501 & 0.195 & 0.838 & 0.95 & 0.983 \\
    Our (resnet50 w/ attention BN) & unrectified & 0.1158 & 0.7714 & 4.865 & 0.175 & 0.861 & 0.961 & 0.988 \\
    \hline
    Our (resnet18 w/ attention IN) & unrectified & 0.1253 & 0.9420 & 5.367  & 0.187 & 0.846 & 0.954 & 0.985 \\
    Our (resnet50 w/ attention IN) & unrectified & 0.1067 & 0.6866 & 4.585  & 0.163 & 0.879 & 0.968 & 0.991 \\
\end{tabular}
}
\caption{Qualitative result comparison on KITTI improved ground truth eigen split. All state of the art methods predict depth from rectified distorted image unlike the proposed method that predict depth capped at 80m from unrectified distorted image. BN: Batch normalization and IN: Instance normalization.}
\label{tab:qual}
\end{table*}

\begin{figure*}[!ht]
  \centering
  \resizebox{\textwidth}{!}{
  \newcommand{\turnheightnew}{0.25\columnwidth}
\centering

\begin{tabular}{@{\hskip 0.5mm}c@{\hskip 0.5mm}c@{\hskip 0.5mm}c@{\hskip 0.5mm}c@{\hskip 0.5mm}c@{}}

{\rotatebox{90}{\hspace{6mm}\shortstack{\Large Unrectified \\ \Large raw input}}} &
\includegraphics[height=\turnheightnew]{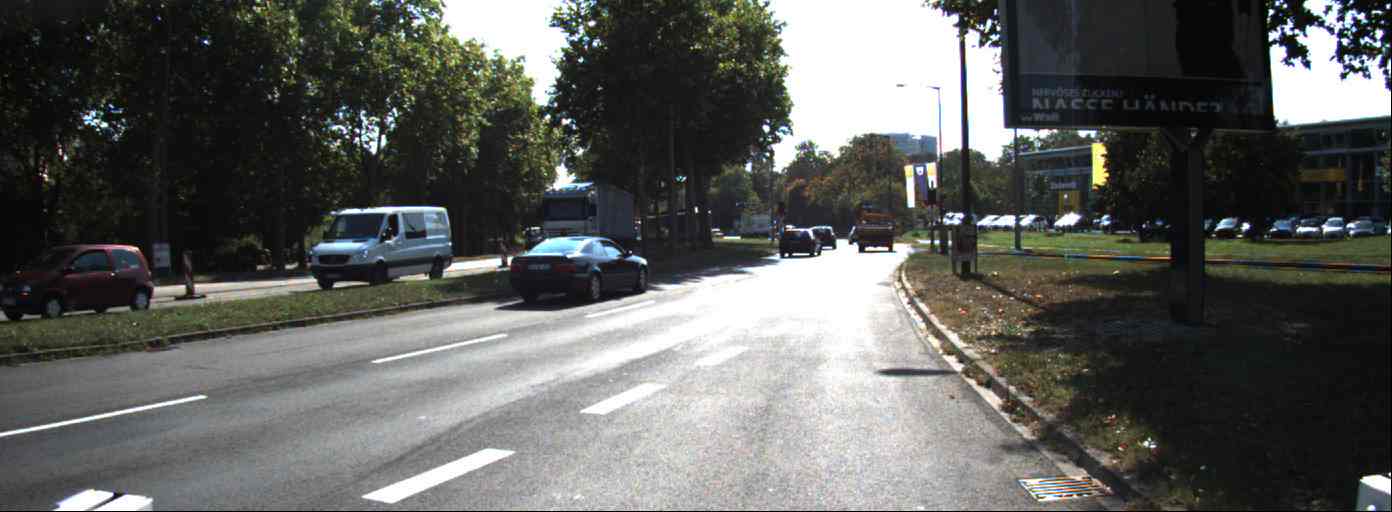} &
\includegraphics[height=\turnheightnew]{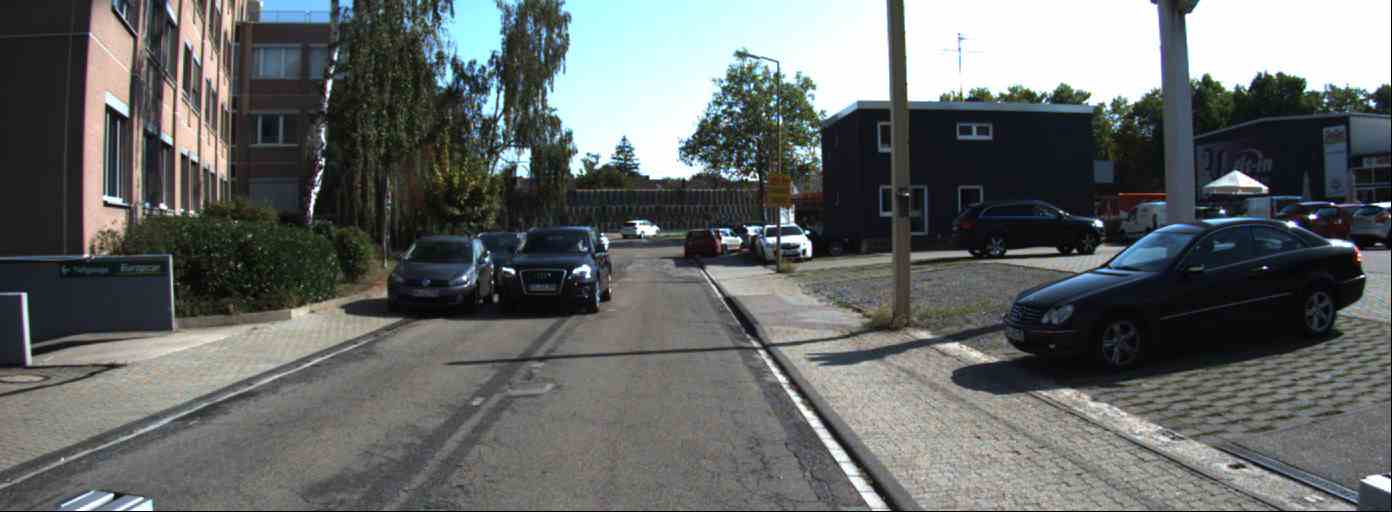} &
\includegraphics[height=\turnheightnew]{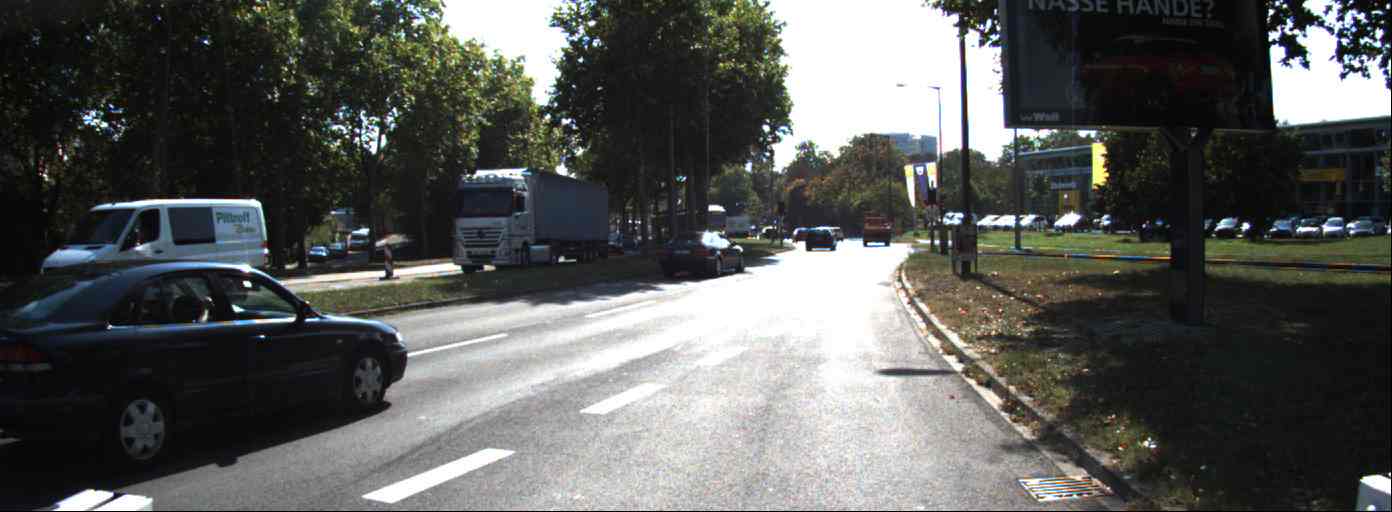} &
\includegraphics[height=\turnheightnew]{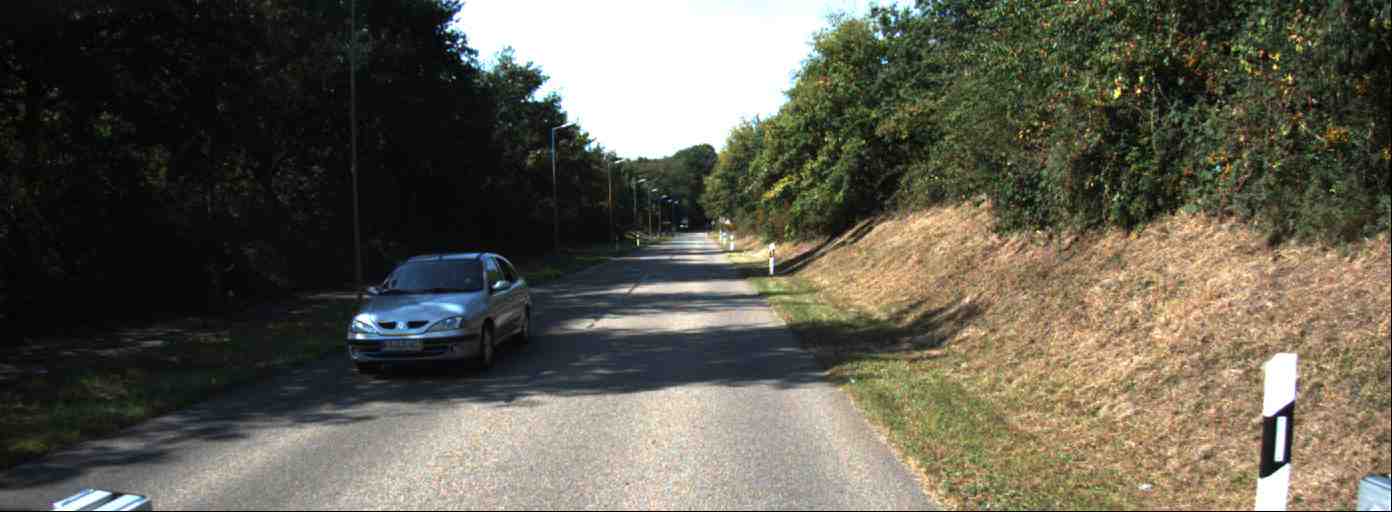}\\

{\rotatebox{90}{\hspace{0mm}\shortstack{\Large LiDAR \\ \Large depth \\\Large groundtruth}}} &
\includegraphics[height=\turnheightnew, width=56mm]{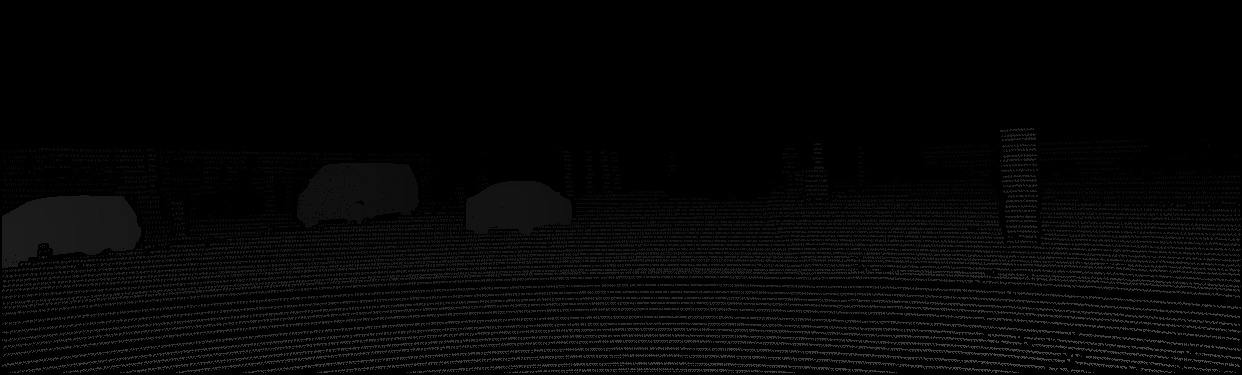} &
\includegraphics[height=\turnheightnew, width=56mm]{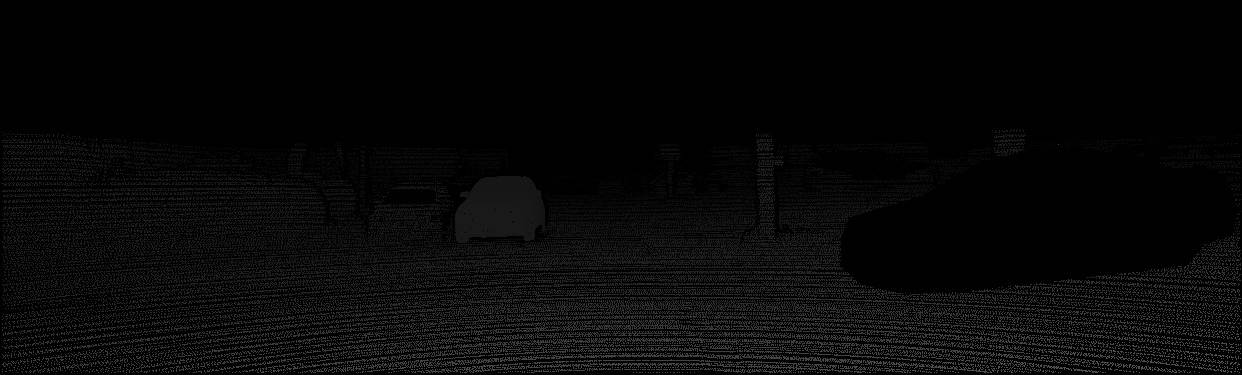} &
\includegraphics[height=\turnheightnew, width=56mm]{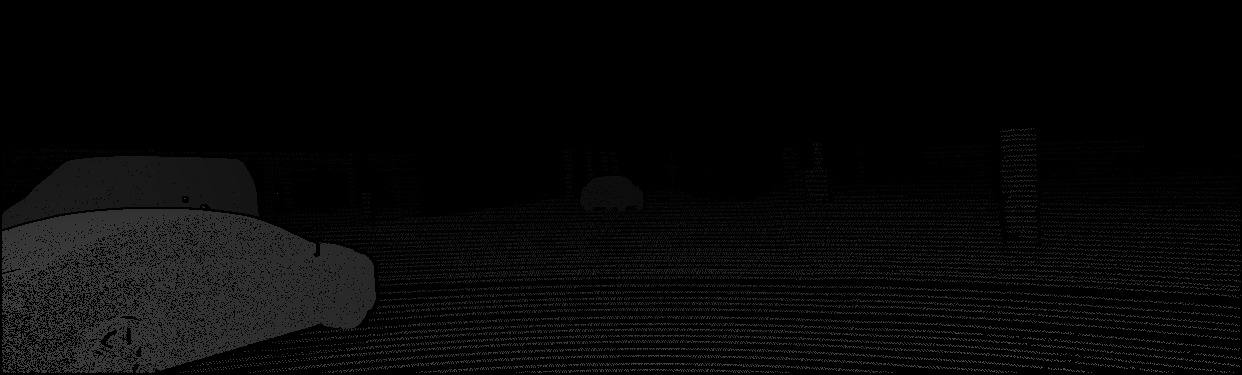} &
\includegraphics[height=\turnheightnew, width=56mm]{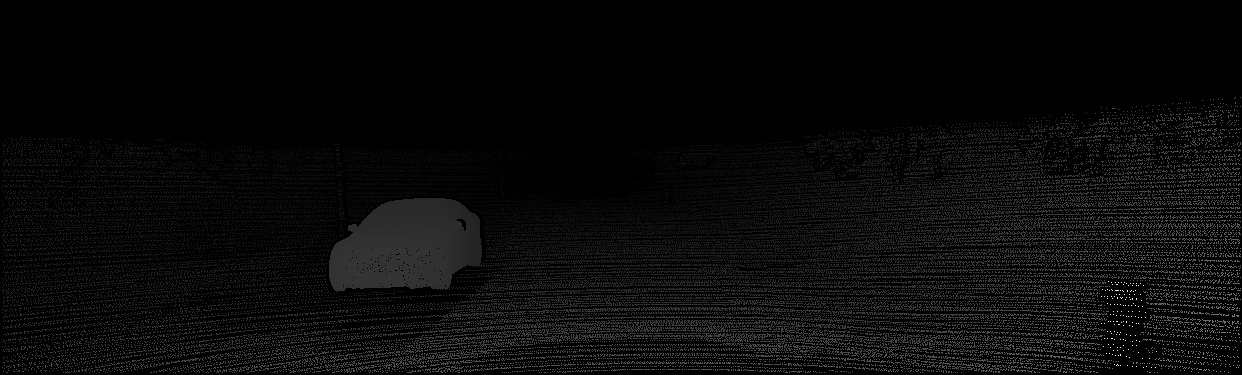}\\

{\rotatebox{90}{\hspace{0mm}\shortstack{\Large Predicted \\ \Large Depth map}}} &
\includegraphics[height=\turnheightnew]{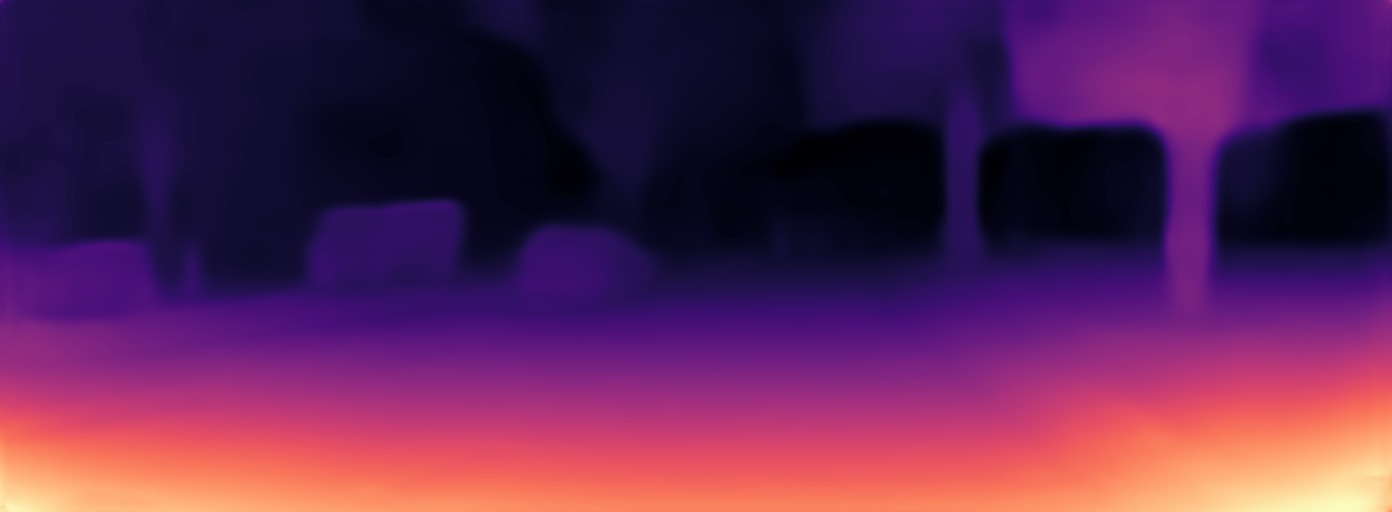} &
\includegraphics[height=\turnheightnew]{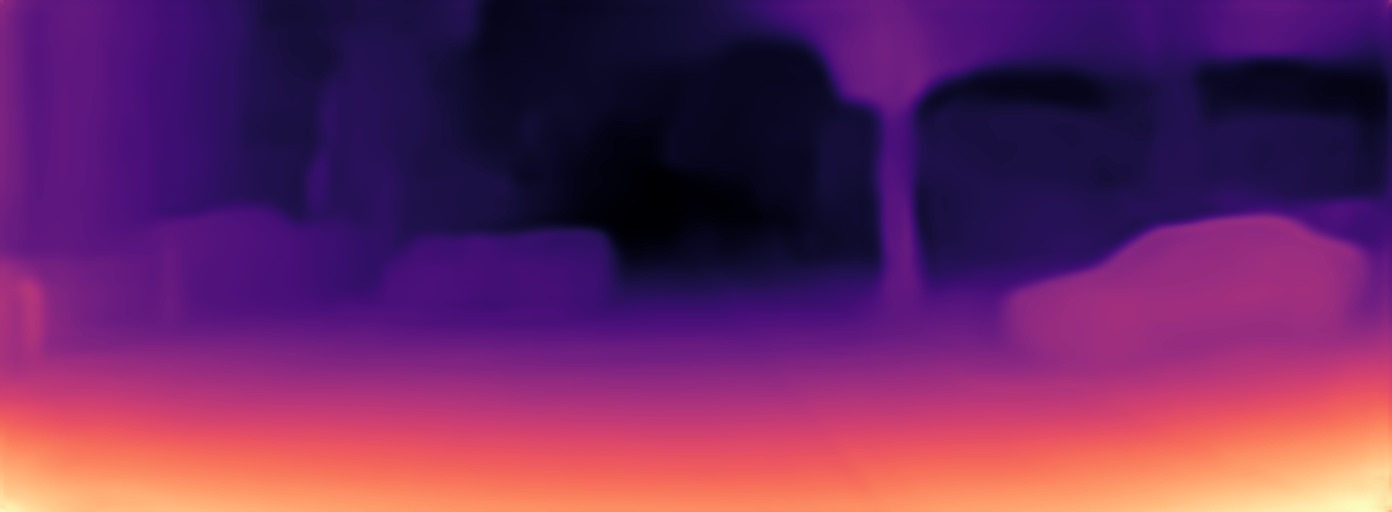} &
\includegraphics[height=\turnheightnew]{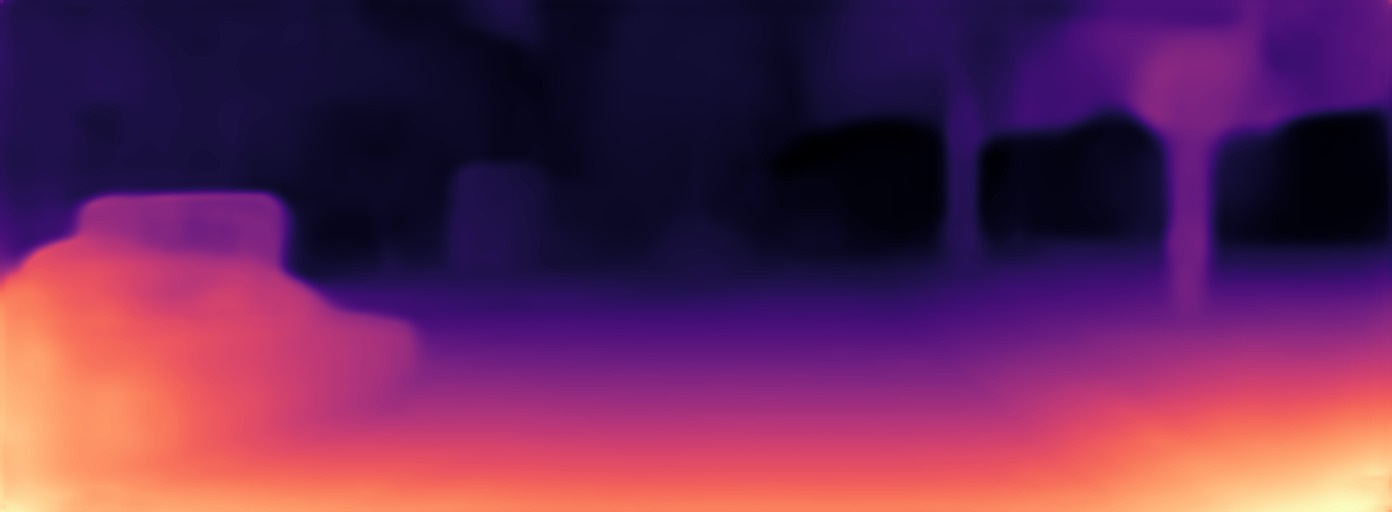} &
\includegraphics[height=\turnheightnew]{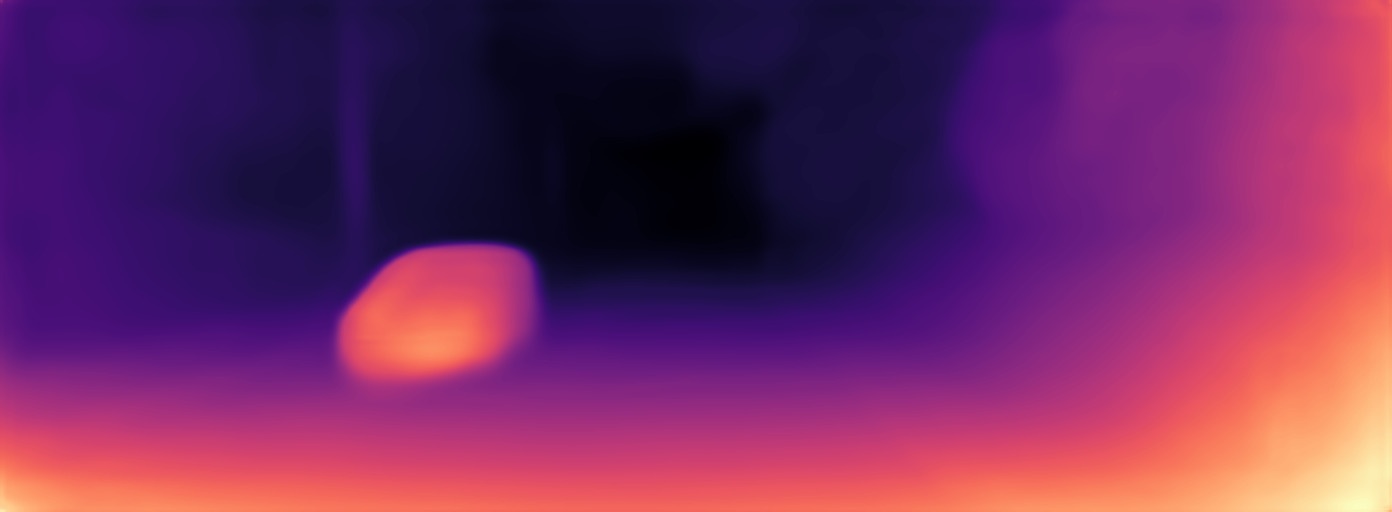} \\

\end{tabular}
}
  \caption{Qualitative result comparison on KITTI unsynced+unrectified   dataset. Figure best viewed in color.}
  \label{fig:qual}
\end{figure*}

\section{Self-Attention network}

Most of the depth estimation networks like  \cite{godard2019digging},\cite{garg2016unsupervised} use convolutions for capturing the local information. But it is well known fact that receptive field of convolutions is quite small. Inspired by \cite{zhang2018self}, we introduce self-attention at different stages within our Resnet encoder \cite{he2016deep} of the depth estimation network to handled long range relations. This enables us to incorporate the contextual information into the high dimensional feature maps. This particularly constrains the network to produce more robust feature maps which result in much improved depth estimation.To the best of knowledge our architecture is the first to incorporate contextual information at different layers of the network for the purpose of unsupervised depth estimation. As a proof of our contribution we visualize attention maps of few samples in Fig.~\ref{fig:att}. We can observe how the network learns to attend pixels belonging to the same attribute or texture for a particular query pixel.

\section{Handling variation in style} 

We believe that major issue in training such models is the high variation in style of the input images which makes it difficult for the model from reaching its best potential. Therefore taking inspiration from \cite{huang2017arbitrary} we normalise feature maps after each convolutional layer. This trivial procedure particularly normalises style of every training image and thus facilitates better training as the model discards the style information and focuses only on the content information to infer depth.

\section{Experiments and results}

We use the KITTI raw (unsynced+unrectified) dataset \cite{geiger2013vision} for training. We extracted 652 of the 697 eigen test split to evaluate unrectified depth with improved KITTI groundtruth depth maps on which all our qualitative results are done, as shown in Table~\ref{tab:qual}.
Our proposed depth estimation method is evaluated using metrics in \cite{eigen2014depth} for a fair comparison with other state of the art methods for rectified images. Note that as there is no previous benchmark for depth estimation on unrectified images, we compare our results with those of rectified images. For evaluating predicted unrectified depthmap, we undistort the unrectified depth map and compare it to the rectified ground truth. The main limitation of proposed method is the transferability to other dataset captured with a different camera. As distortion varies from one camera to another, direct transfer learning would be difficult, but fine-tuning is the best way until and unless the projection doesn't break. The projection used in the propose method will fail with extreme distortion like wide-angle lens.

\section{Conclusion}

We proposed a novel self-attention network for learning monocular depth from unrectified video frames. Our attention framework is able to capture long distance contextual information which results in much sharper depth maps. We also handle style variance in the training distribution by using instance normalisation. Finally, we evaluated the unrectified depth estimation against state of the art methods on rectified depth estimation and achieved comparable results.
	
\bibliography{refs}


\clearpage
\onecolumn
\begin{center}
	{\Large Supplementary Material:\\ 
		Self-Attention Dense Depth Estimation Network for Unrectified Video Sequences}
\end{center}

\begin{figure*}[!ht]
	\centering
	\resizebox{0.85\textwidth}{!}{
		\newcommand{\turnheightnew}{0.25\columnwidth}
\centering

\begin{tabular}{@{\hskip 0.5mm}c@{\hskip 0.5mm}c@{\hskip 0.5mm}c@{\hskip 0.5mm}c@{\hskip 0.5mm}c@{}}

{\hspace{6mm} {\Large Unrectified input image}} &
{\hspace{6mm} {\Large Attention map}} \\

\includegraphics[height=\turnheightnew]{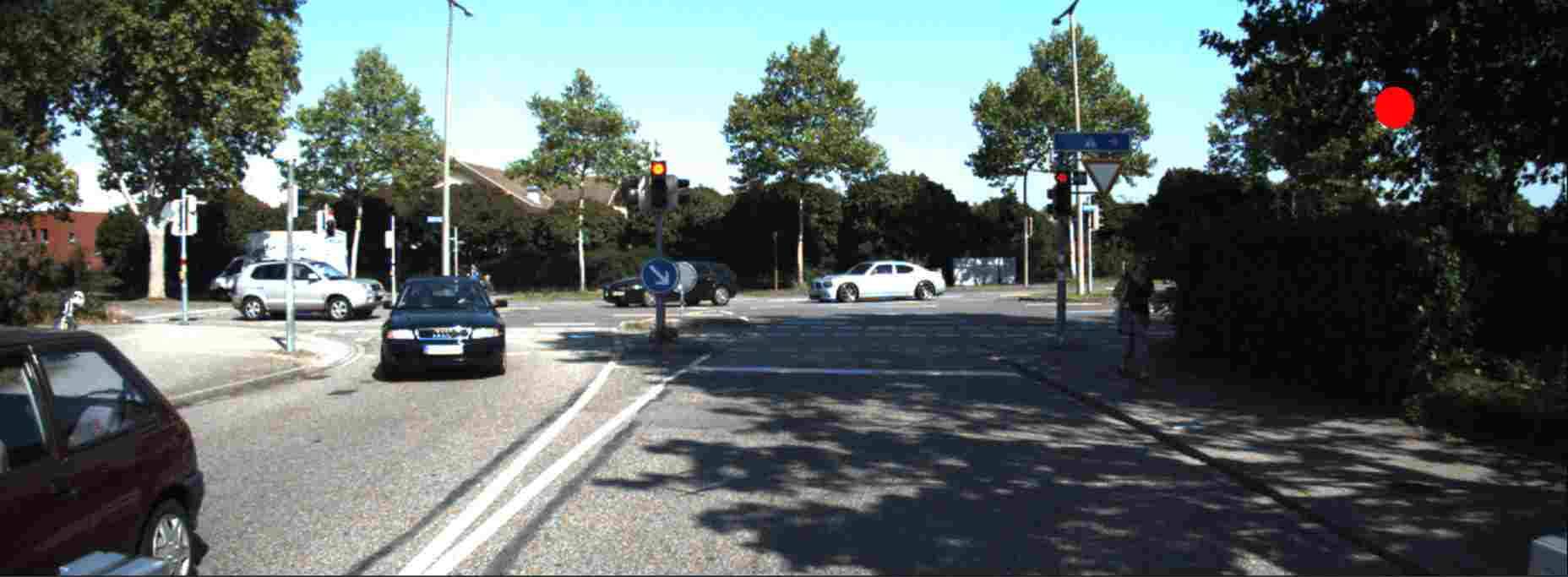} &
\includegraphics[height=\turnheightnew]{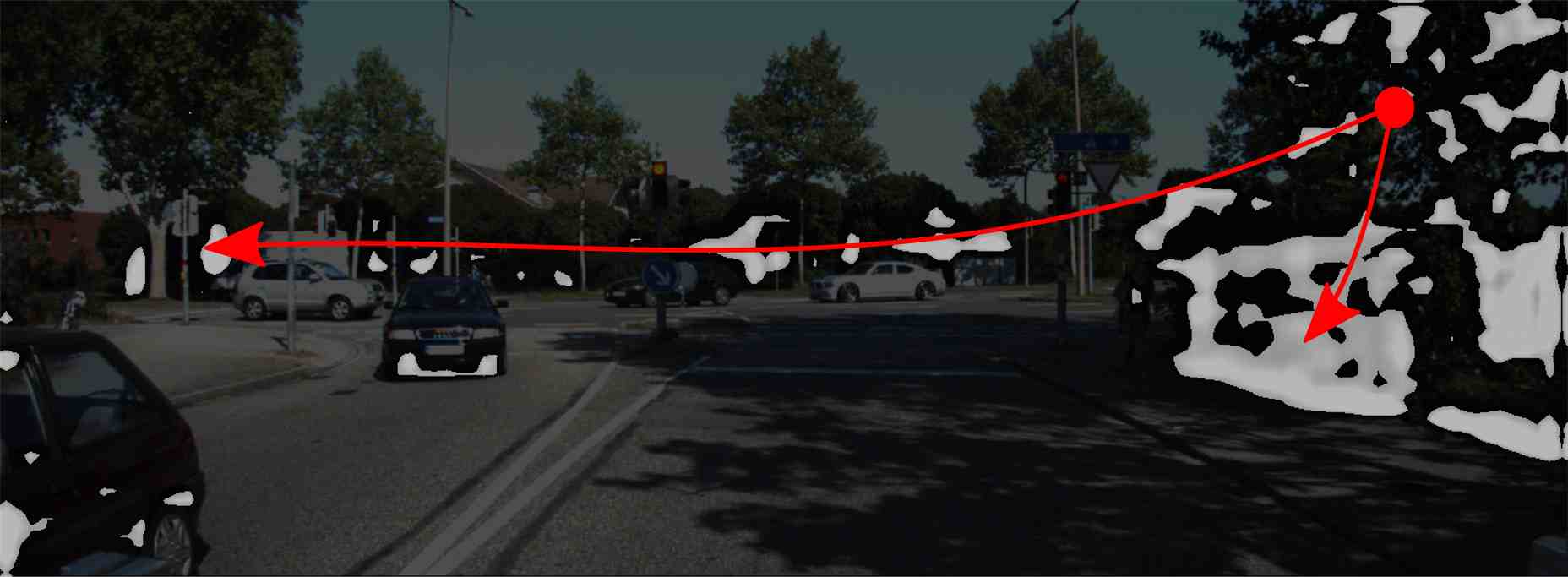} \\

\includegraphics[height=\turnheightnew]{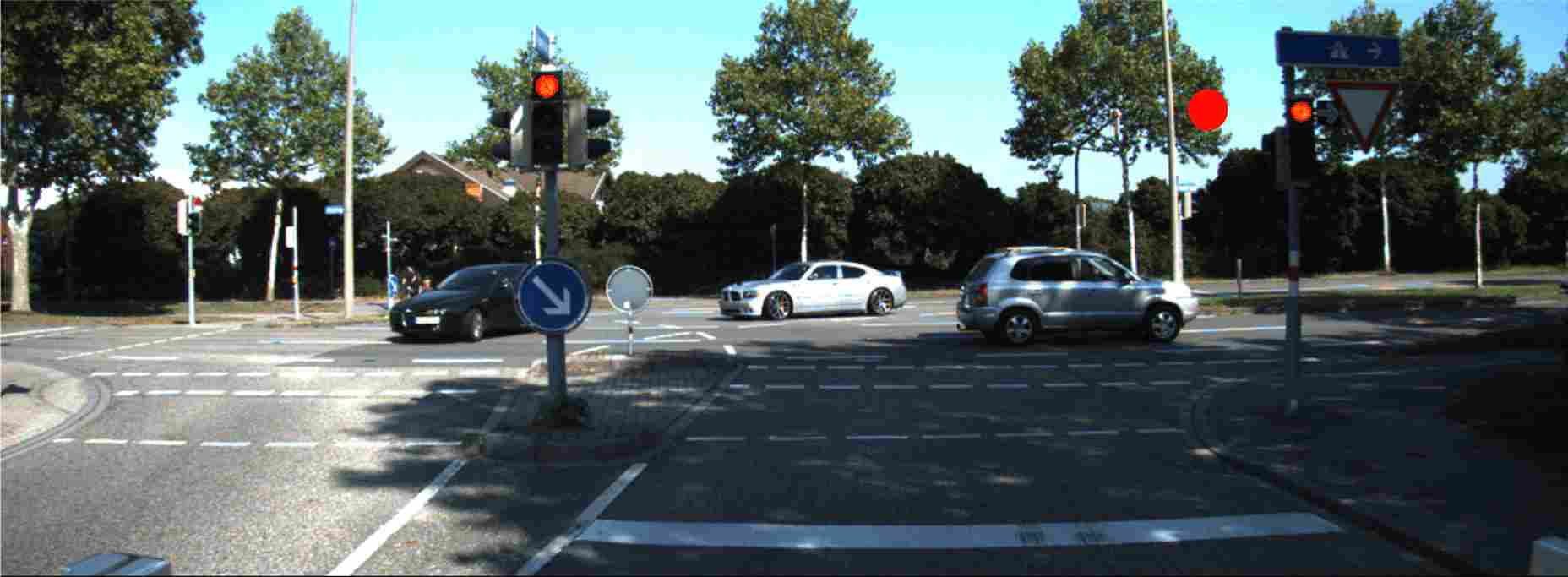} &
\includegraphics[height=\turnheightnew]{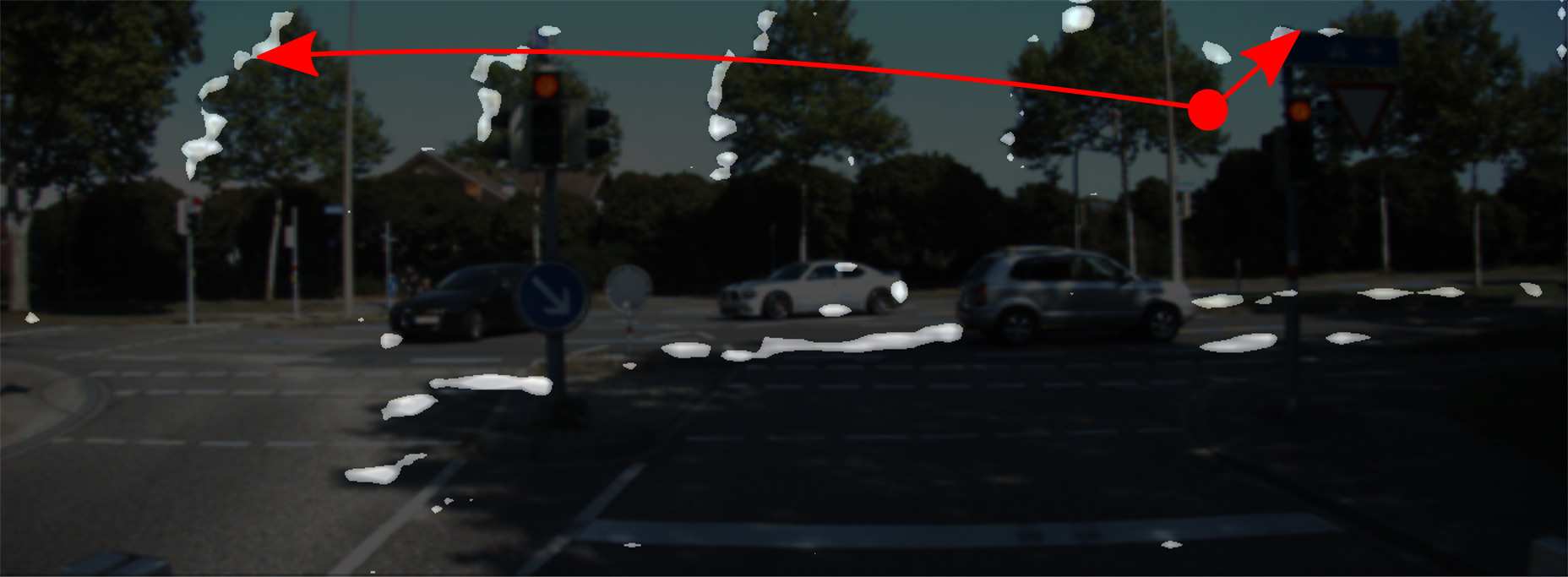} \\

\includegraphics[height=\turnheightnew]{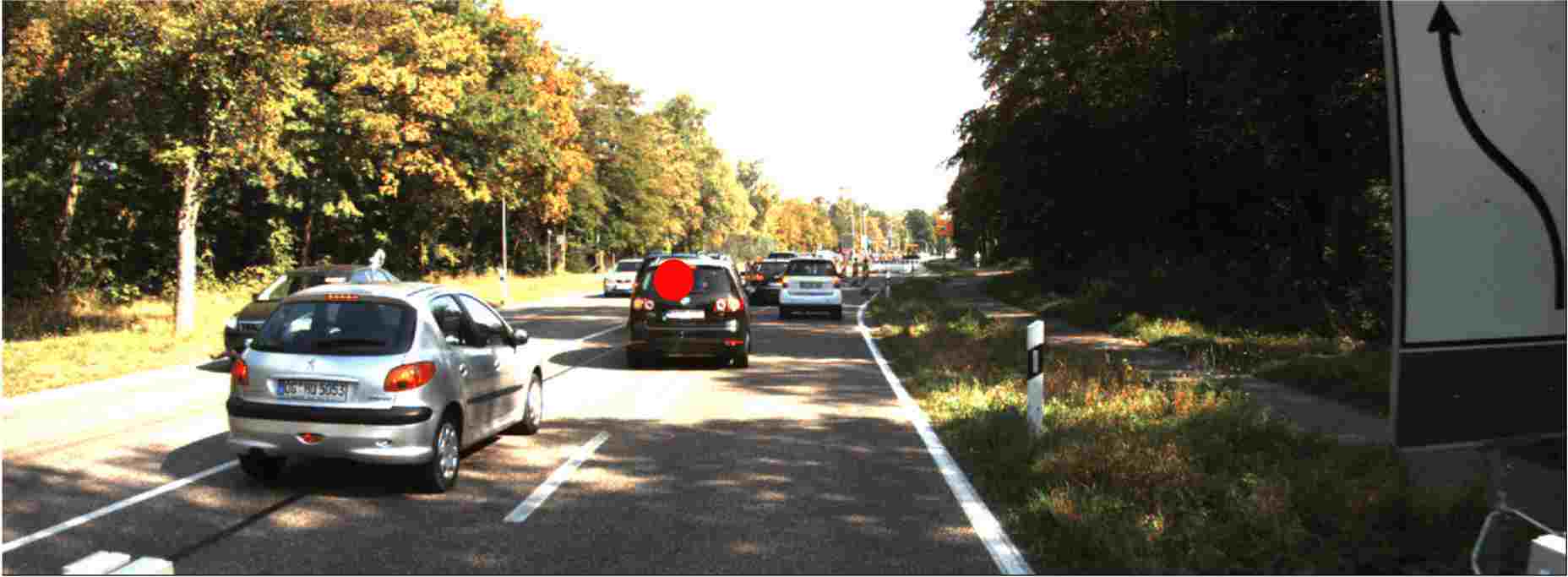} &
\includegraphics[height=\turnheightnew]{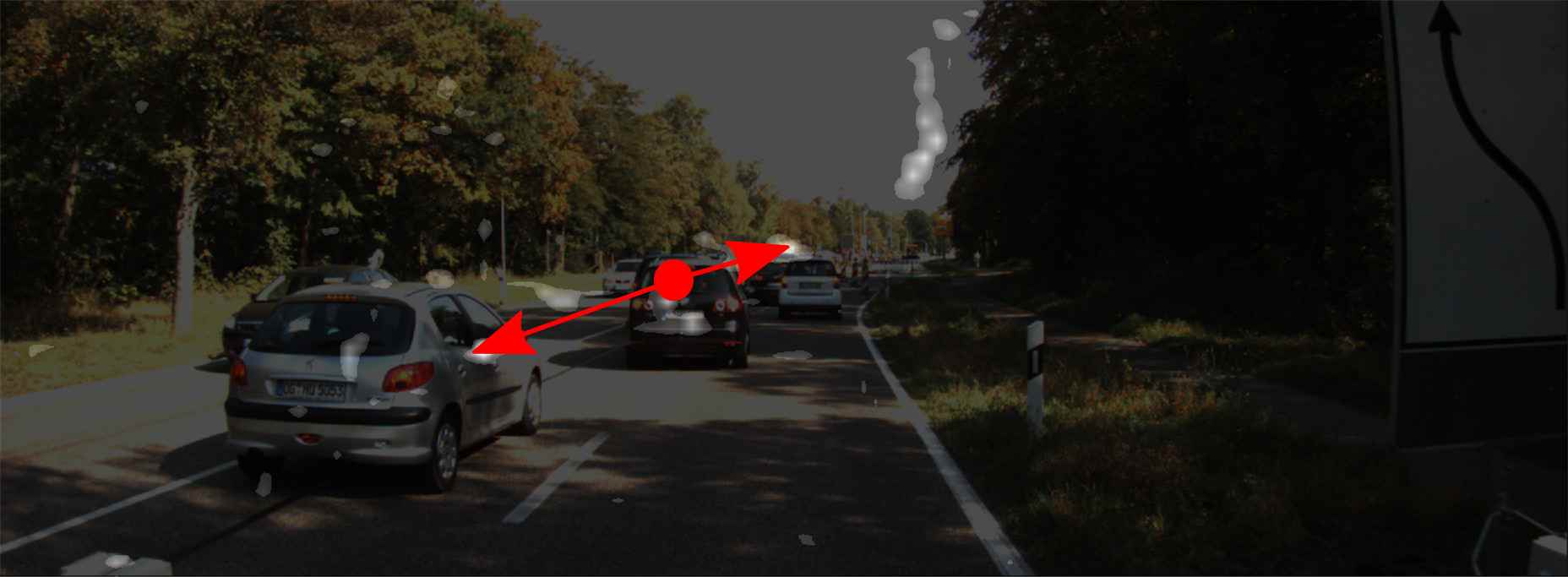} \\

\end{tabular}}
	\caption{Visualization of self-attention map. Left: Shows Query locations marked with a red dot on the unrectified input image. Right: Attention map for those query locations overlayed on the unrectified input image with arrows summarizing most-attended regions.
	}
	\label{fig:att_suppl}
\end{figure*}

\begin{figure*}[!ht]
  \centering
  \resizebox{\textwidth}{!}{
  \newcommand{\turnheightnew}{0.25\columnwidth}
\centering

\begin{tabular}{@{\hskip 0.5mm}c@{\hskip 0.5mm}c@{\hskip 0.5mm}c@{}}

{\hspace{6mm}\shortstack{\Large Unrectified raw input}} &
{\hspace{6mm}\shortstack{\Large LiDAR depth groundtruth}} &
{\hspace{6mm}\shortstack{\Large Predicted Depth map}} \\

\includegraphics[height=\turnheightnew]{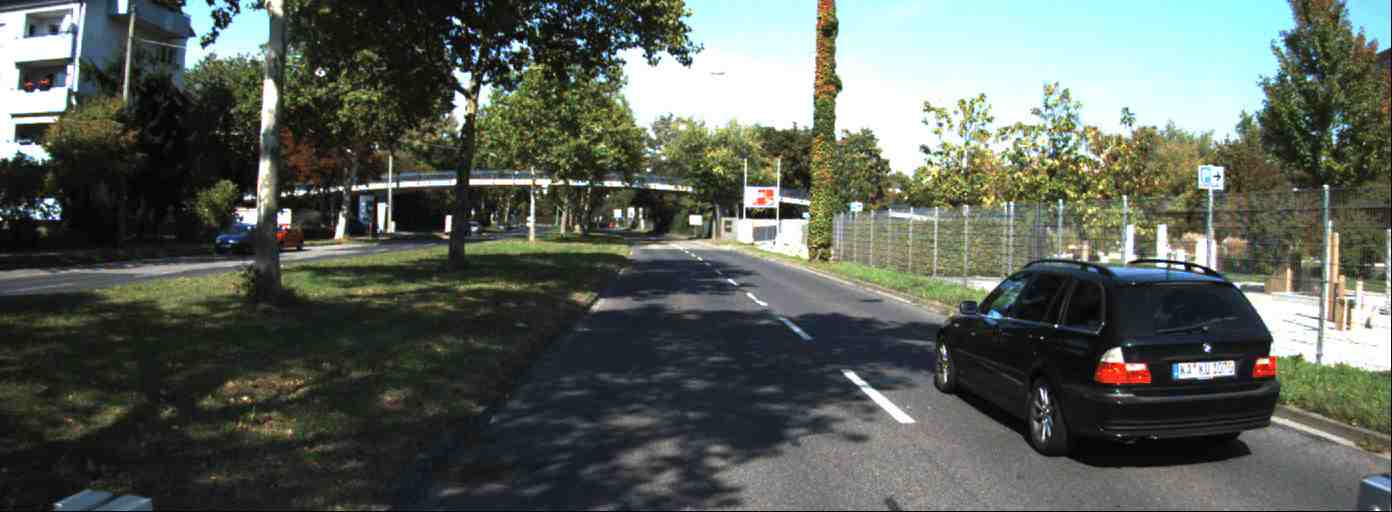} &
\includegraphics[height=\turnheightnew]{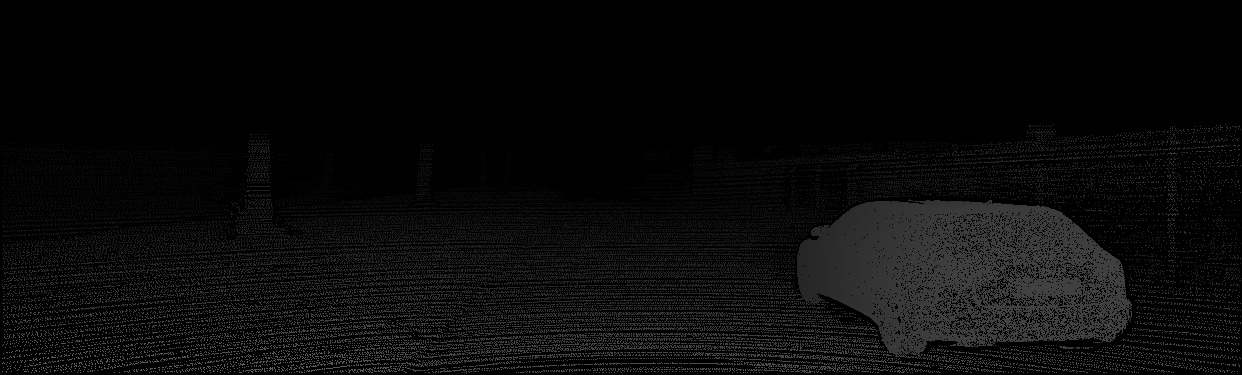} &
\includegraphics[height=\turnheightnew]{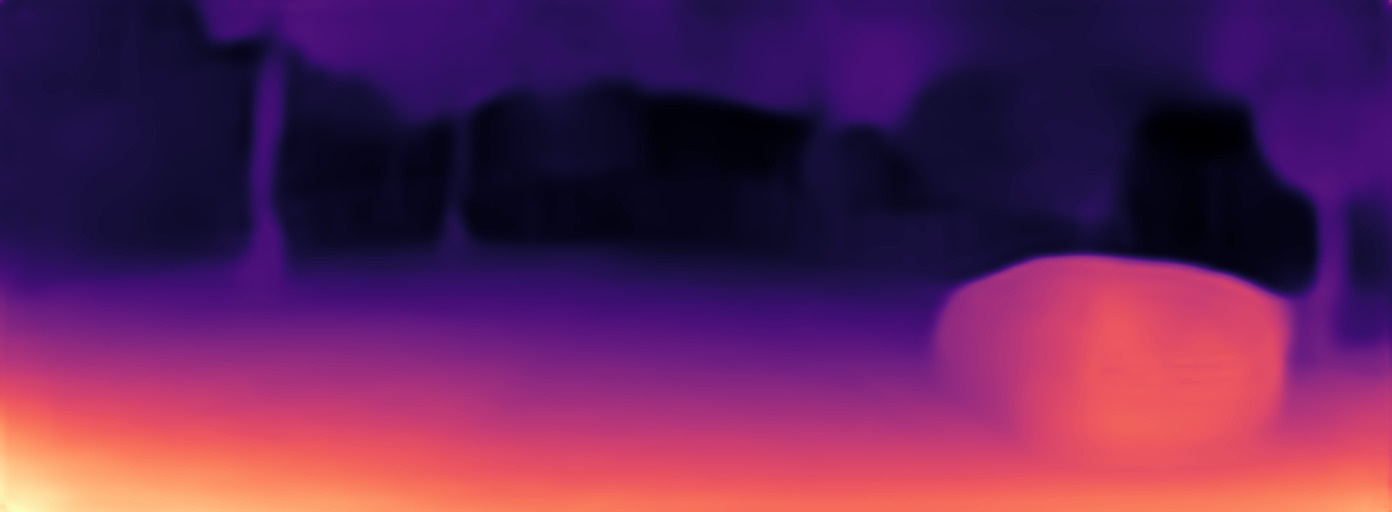}\\

\includegraphics[height=\turnheightnew]{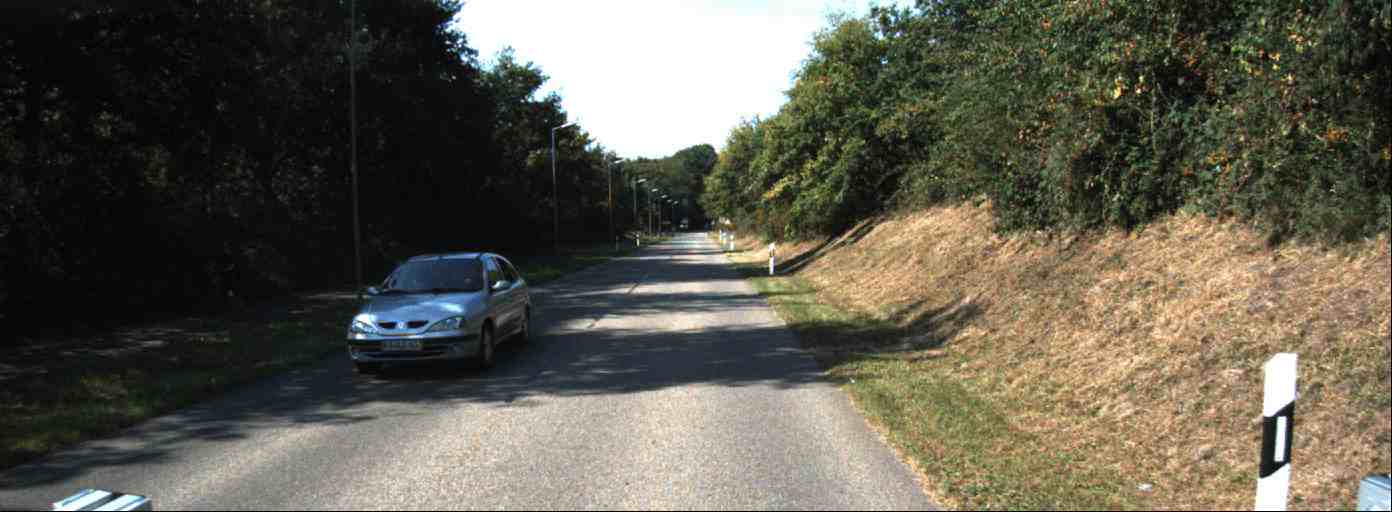} &
\includegraphics[height=\turnheightnew]{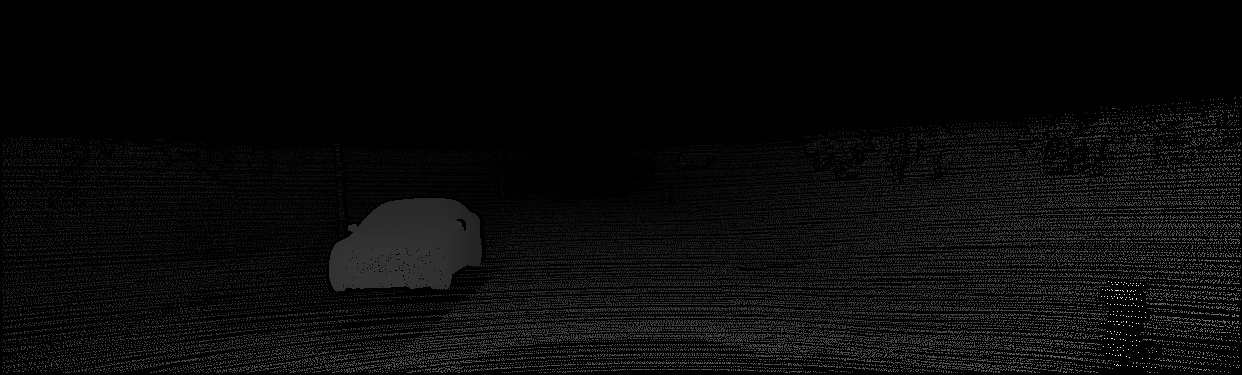} &
\includegraphics[height=\turnheightnew]{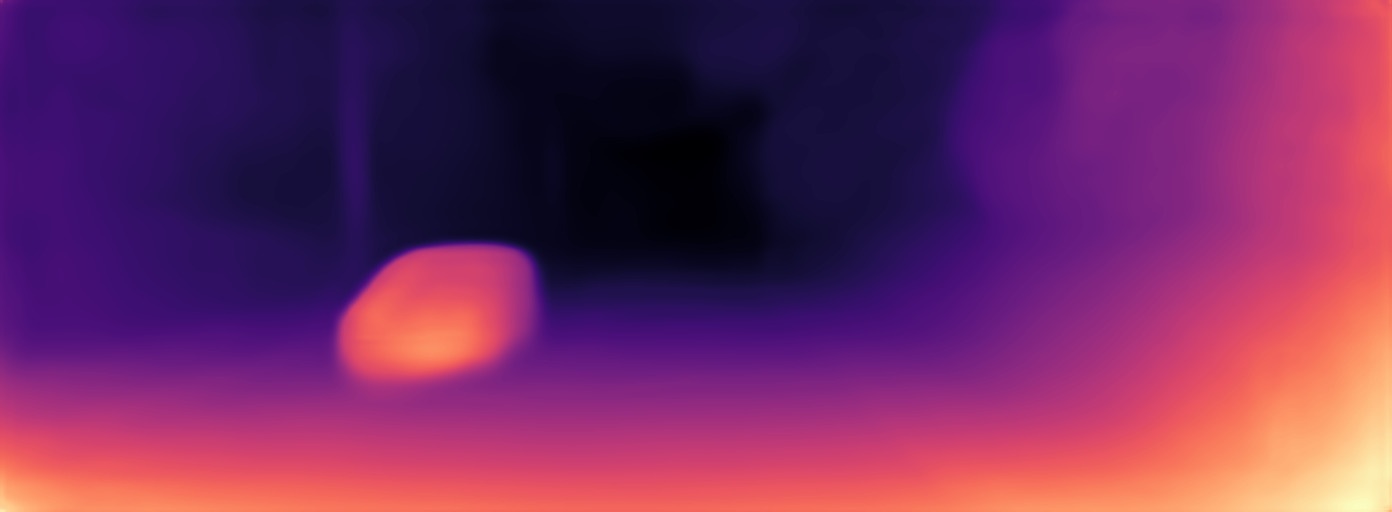}\\

\includegraphics[height=\turnheightnew]{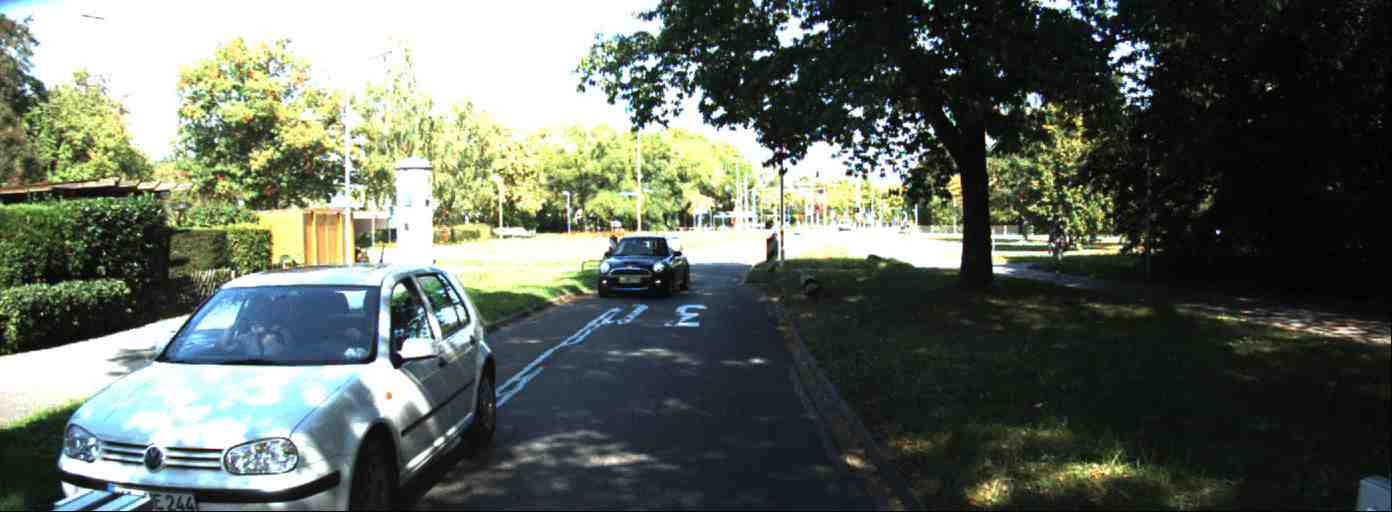} &
\includegraphics[height=\turnheightnew]{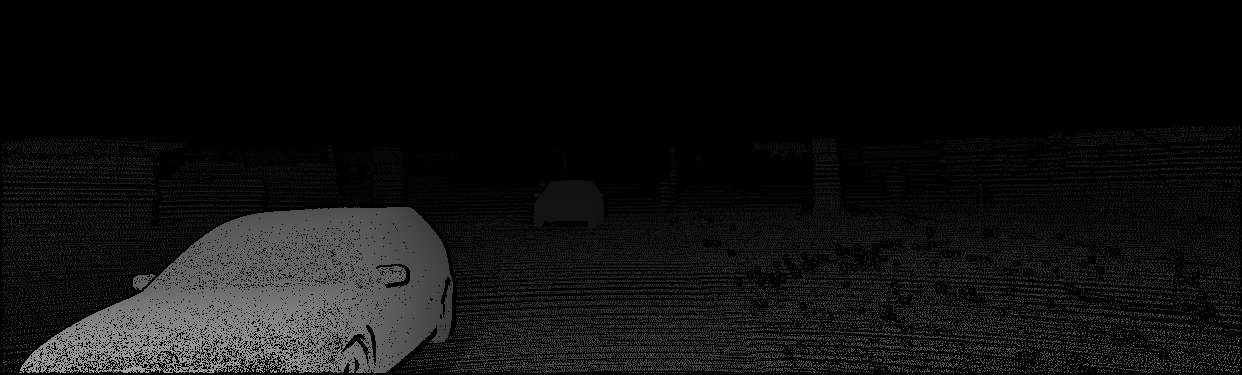} &
\includegraphics[height=\turnheightnew]{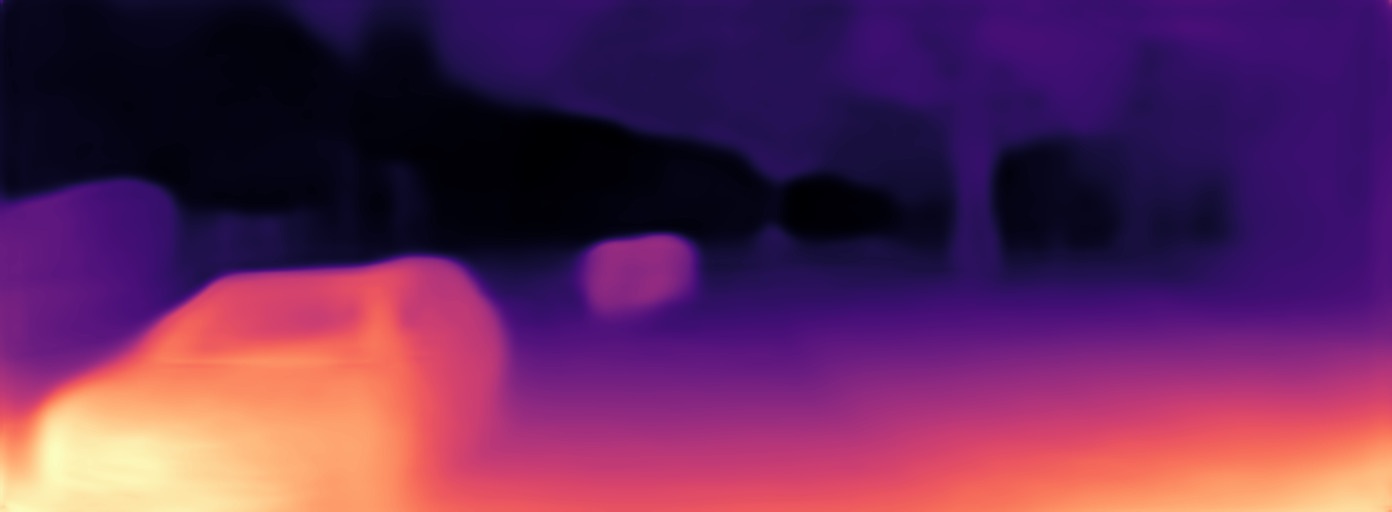}\\

\includegraphics[height=\turnheightnew]{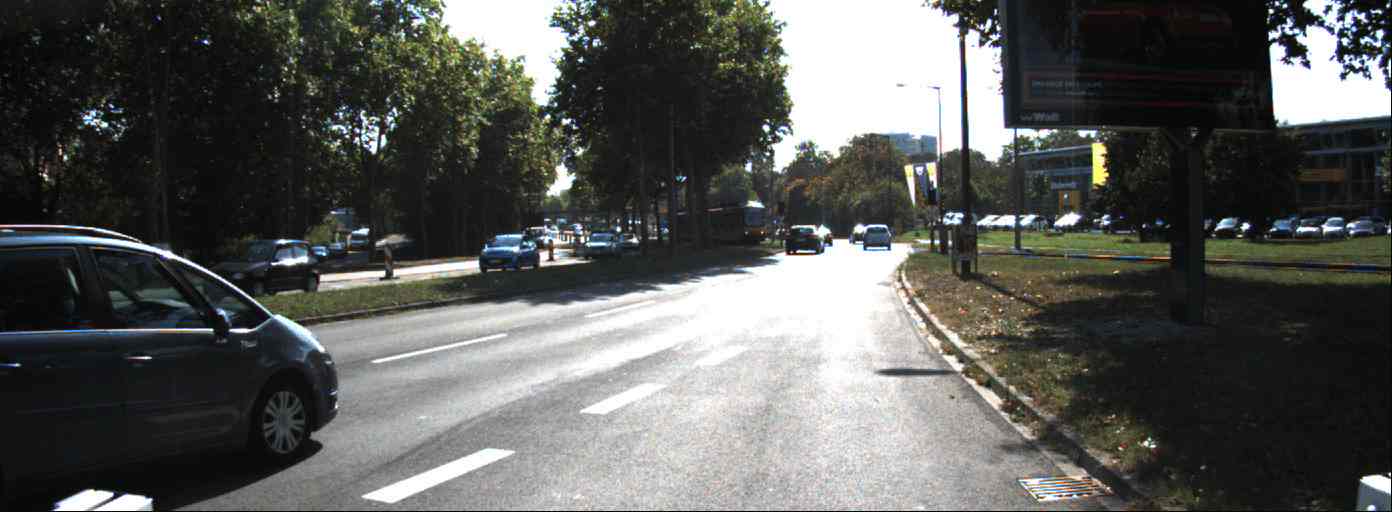} &
\includegraphics[height=\turnheightnew]{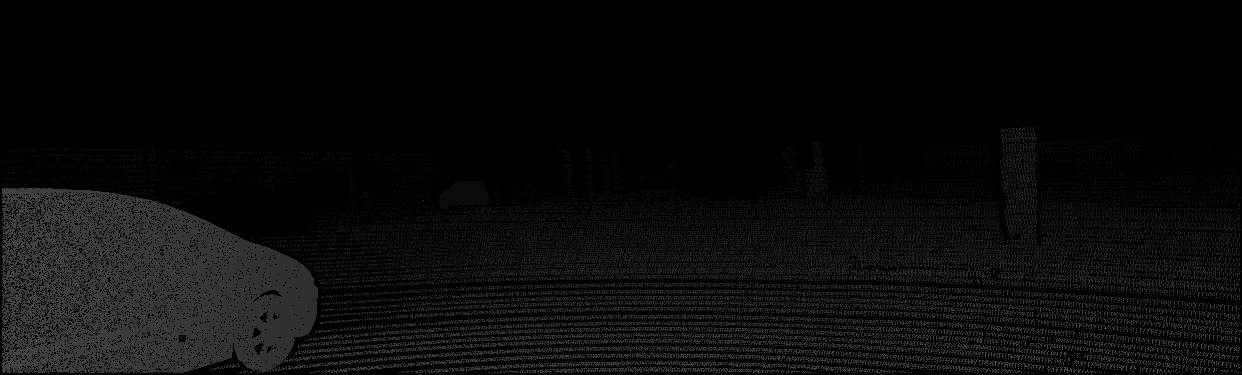} &
\includegraphics[height=\turnheightnew]{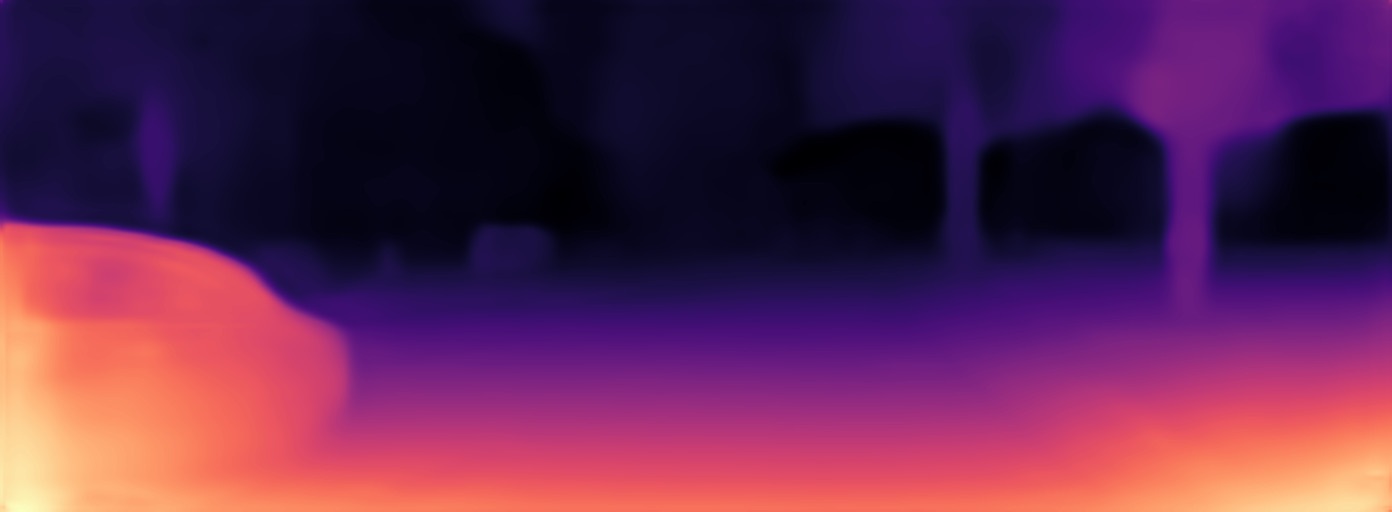}\\

\includegraphics[height=\turnheightnew]{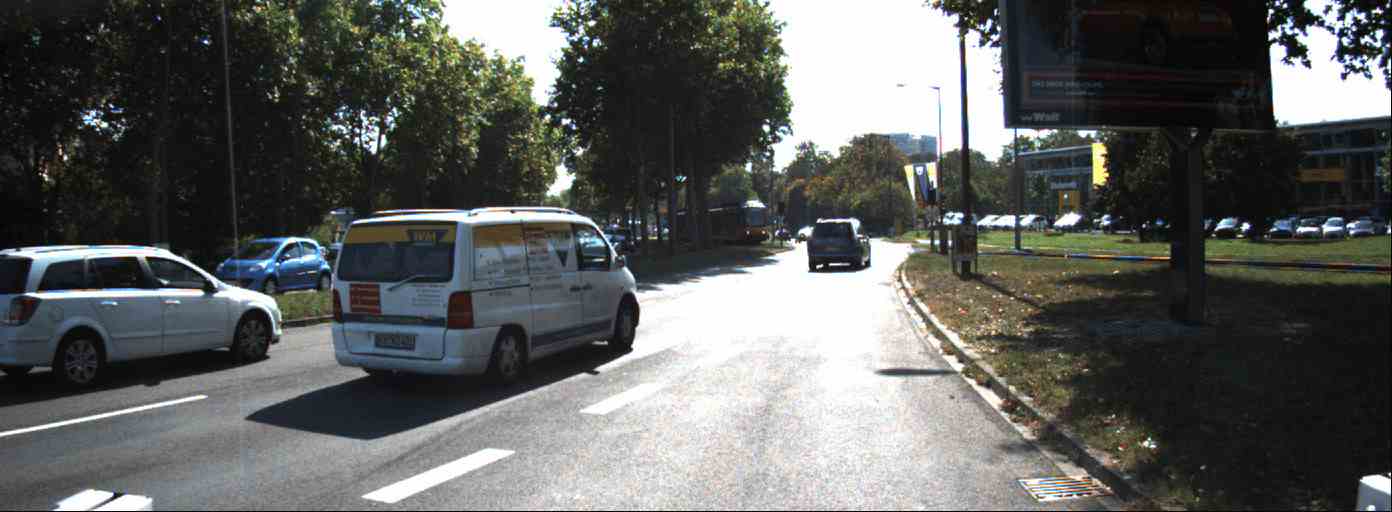} &
\includegraphics[height=\turnheightnew]{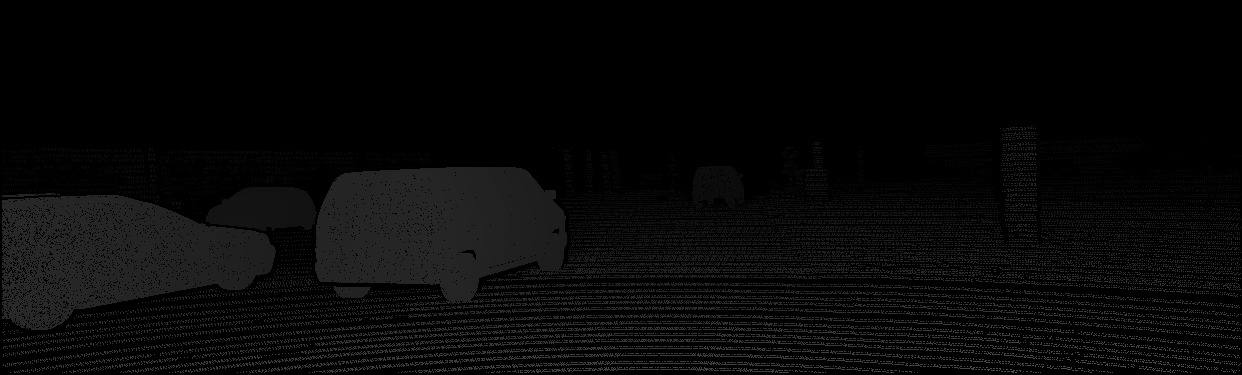} &
\includegraphics[height=\turnheightnew]{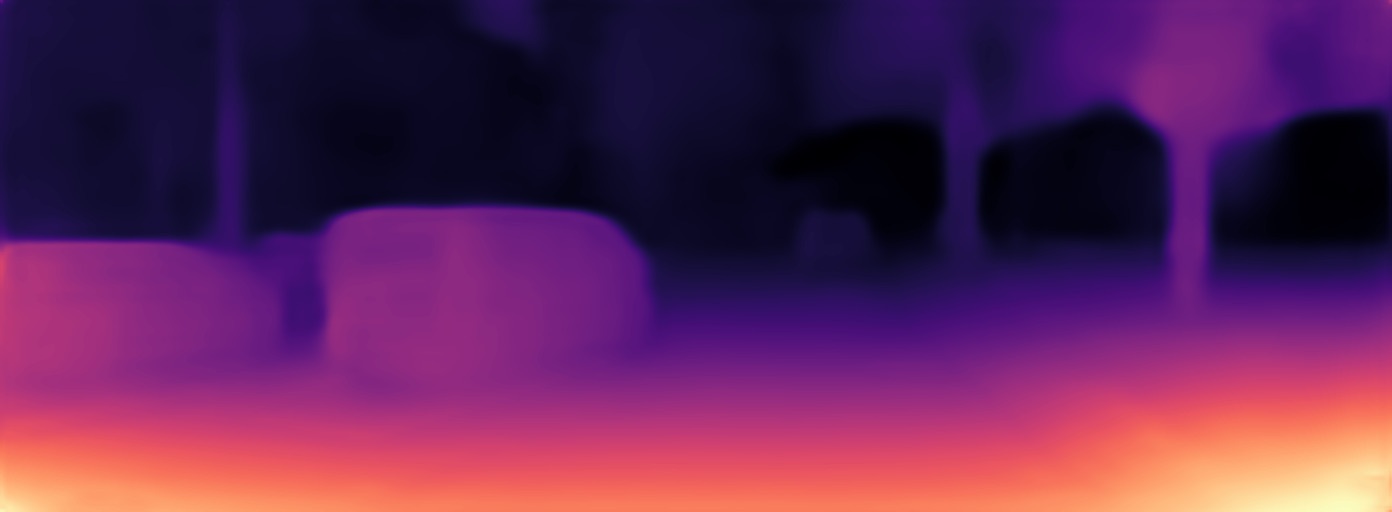}\\

\includegraphics[height=\turnheightnew]{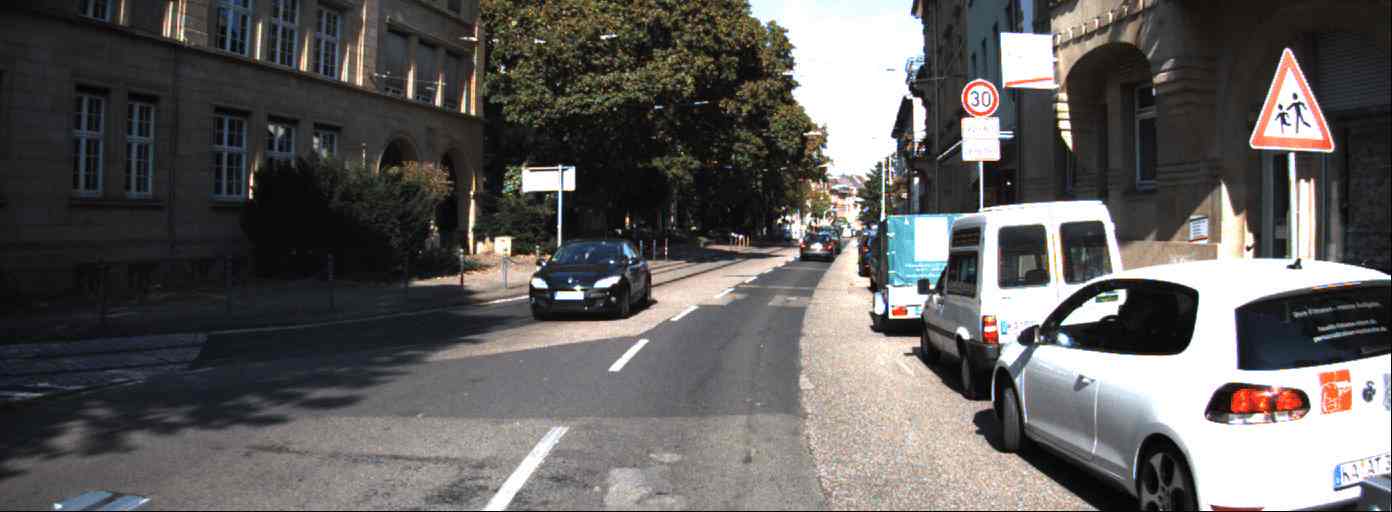} &
\includegraphics[height=\turnheightnew]{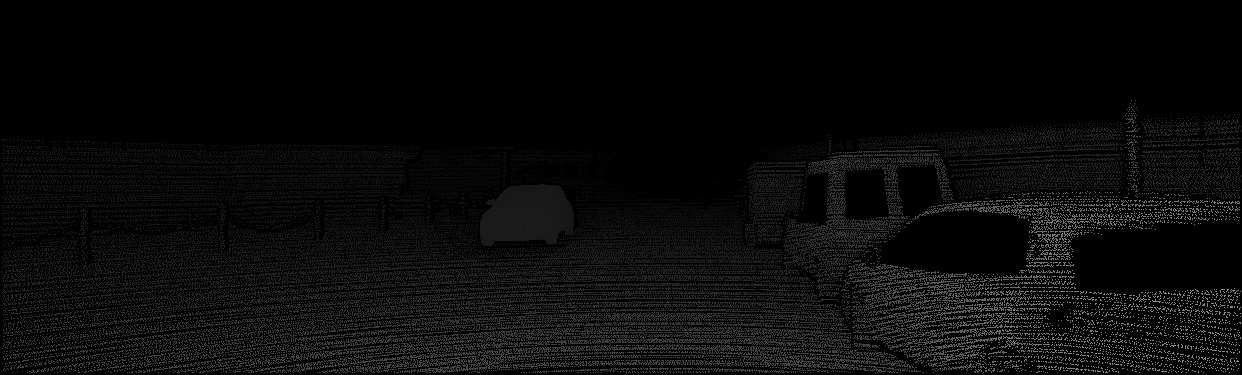} &
\includegraphics[height=\turnheightnew]{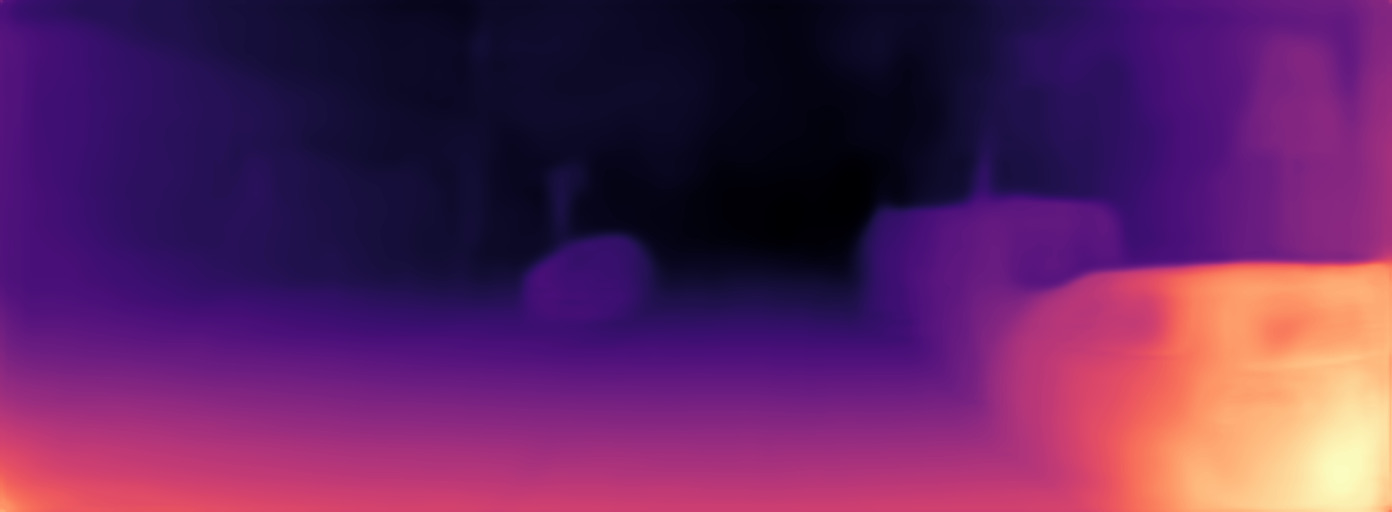}\\

\end{tabular}
}
  \caption{Qualitative result comparison on KITTI unsynced+unrectified dataset. Figure best viewed in color.}
  \label{fig:qual_suppl}
\end{figure*}
	
\end{document}